\documentclass[10pt,twocolumn,letterpaper]{article}

\usepackage[pagenumbers]{cvpr} % To force page numbers, e.g. for an arXiv version

\usepackage[dvipsnames]{xcolor}

\usepackage{graphicx}
\usepackage{amsmath}
\usepackage{amssymb}
\usepackage{booktabs}
\usepackage{makecell}

\usepackage{etoc}
\usepackage{titletoc}
\usepackage{wrapfig}
\usepackage{multirow}
\usepackage{cuted}
\usepackage{caption, subcaption, overpic, textpos}
\usepackage{enumitem}
\usepackage{color, colortbl}
\usepackage{comment}
\definecolor{LightCyan}{rgb}{0.88,1,1}
\definecolor{lightgray}{gray}{.9}

\newcommand{\algname}{ViDAR\xspace}
\newcommand{\encoder}{History Encoder\xspace}
\newcommand{\render}{Latent Rendering\xspace}
\newcommand{\decoder}{Future Decoder\xspace}

\newcommand{\myparagraph}[1]{\vspace{3pt}\noindent{\bf #1}}

\usepackage{lipsum}
\usepackage{bbding}
\usepackage{diagbox}
\usepackage{algorithm}
\usepackage{listings}
\usepackage{etoolbox}
\makeatletter
\AfterEndEnvironment{algorithm}{\let\@algcomment\relax}
\AtEndEnvironment{algorithm}{\kern2pt\hrule\relax\vskip3pt\@algcomment}
\let\@algcomment\relax
\newcommand\algcomment[1]{\def\@algcomment{\footnotesize#1}}
\renewcommand\fs@ruled{\def\@fs@cfont{\bfseries}\let\@fs@capt\floatc@ruled
  \def\@fs@pre{\hrule height.8pt depth0pt \kern2pt}%
  \def\@fs@post{}%
  \def\@fs@mid{\kern2pt\hrule\kern2pt}%
  \let\@fs@iftopcapt\iftrue}
\makeatother

\definecolor{cvprblue}{rgb}{0.21,0.49,0.74}
\usepackage[pagebackref=true,breaklinks,colorlinks,citecolor=cvprblue]{hyperref}

\title{
Visual Point Cloud Forecasting enables Scalable Autonomous Driving
}

\author{
{Zetong Yang}
\quad
{Li Chen}
\quad
Yanan Sun
\quad
Hongyang Li \\
[2mm]
OpenDriveLab 
and Shanghai AI Lab 
\\
[1mm]
\normalsize{
\url{https://github.com/OpenDriveLab/ViDAR}} 
\\
[2mm]
\textit{
\textcolor{gray}{In memoriam Prof. Xiao'ou Tang}}
}
\begin{document}

\twocolumn[{
\renewcommand\twocolumn[1][]{#1}%
\maketitle
\begin{center}
    \vspace{-15pt}
    \centering
    \includegraphics[width=\textwidth]{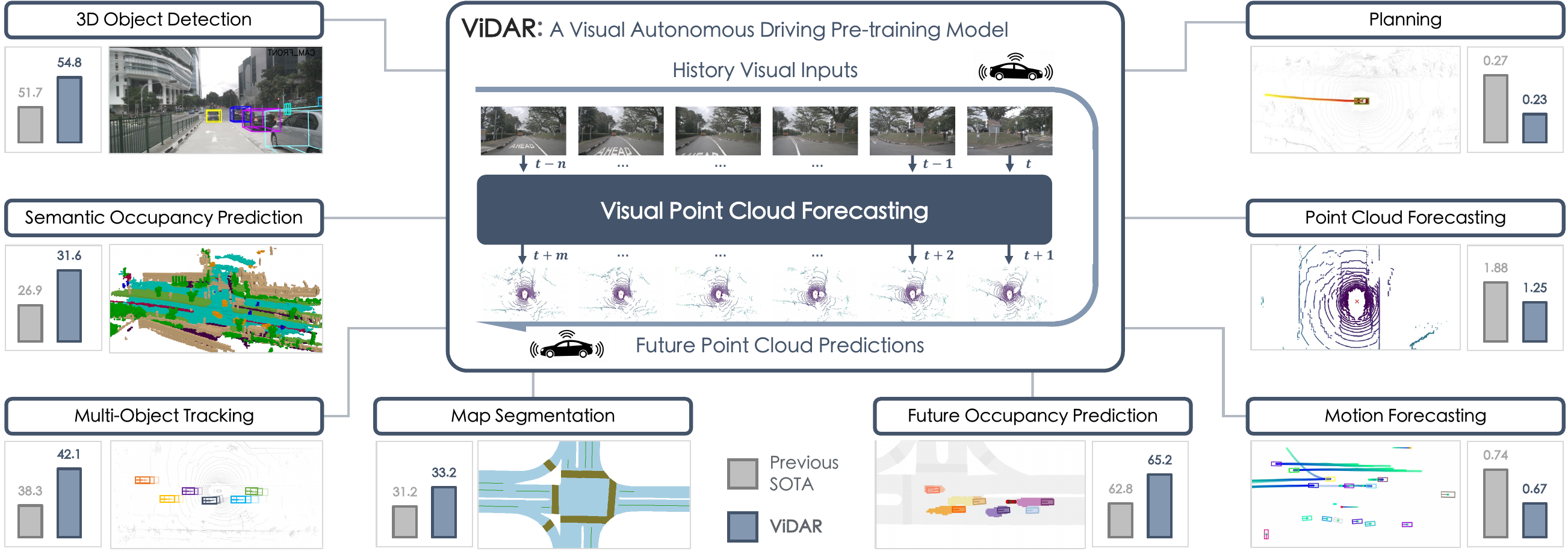}
    \captionof{figure}{\label{fig:teaser}
    \textbf{\algname} is a visual autonomous driving pre-training framework, which leverages the estimation of future point clouds from historical visual inputs as the pre-text task. We term this new pre-text task as \textit{visual point cloud forecasting}. With the aid of \algname, we achieve substantial improvement spanning a diverse spectrum of downstream applications for perception, prediction, and planning.
    }
\end{center}
\vspace{5mm}
}]

\begin{abstract}
In contrast to extensive studies on general vision, pre-training for scalable visual autonomous driving remains seldom explored. Visual autonomous driving applications require features encompassing semantics, 3D geometry, and temporal information simultaneously for joint perception, prediction, and planning, posing dramatic challenges for pre-training. To resolve this, we bring up a new pre-training task termed as \textit{visual point cloud forecasting} -  predicting future point clouds from historical visual input. The key merit of this task captures the synergic learning of semantics, 3D structures, and temporal dynamics. Hence it shows superiority in various downstream tasks. To cope with this new problem, we present \textbf{\algname}, a general model to pre-train downstream visual encoders. It first extracts historical embeddings by the encoder. These representations are then transformed to 3D geometric space via a novel {\render} operator for future point cloud prediction. Experiments show significant gain in downstream tasks, e.g., $3.1$\% NDS on 3D detection, $\sim$10\% error reduction on motion forecasting, and $\sim$15\% less collision rate on planning.
\end{abstract}

\vspace{-.5cm}
\section{Introduction}
\label{sec:intro}
Recently, the community has witnessed rapid development in visual, or camera-only autonomous driving, with input being monocular or multi-view images \cite{wang2021fcos3d,wang2022mvfcos3d++,liu2023sparsebev,wu2022trajectoryguided,sima2023drivelm}. Leveraging visual inputs alone, existing approaches demonstrate superior capability of extracting Bird's-Eye-View (BEV) features \cite{li2022bevformer,yang2023bevformerV2,liu2022petr,CaDDN,huang2021bevdet,chen2022persformer}, and performing well in perception \cite{liu2023bevfusion,li2022uvtr,Tong_2023_ICCV,huang2023tri,wei2023surroundocc}, prediction \cite{vip3d,ijcai2022p785,hu2023_uniad}, and planning \cite{jia2023thinktwice,jia2023driveadapter}. 
Despite significant improvements in applications, these models rely on precise 3D annotations to a great extent, which are often difficult to collect, \eg, semantic occupancy \cite{behley2019iccv,openscene2023}, 3D boxes \cite{KITTIDATASET1,Waymo,nuscenes2019}, trajectories \cite{waymomotion}, and thus are challenging to scale up for production.

Considering the expensive labeling workflow, pre-training \cite{gui2023survey,balestriero2023cookbook,shwartzziv2023compress} has emerged as a crucial approach to scale up downstream applications. The key idea is to define pretext tasks that leverage large amounts of readily available data to learn meaningful representations. This enhances downstream performance though labeled data is limited.

\begin{figure}[t]
  \centering
  \includegraphics[width=1.0\linewidth]{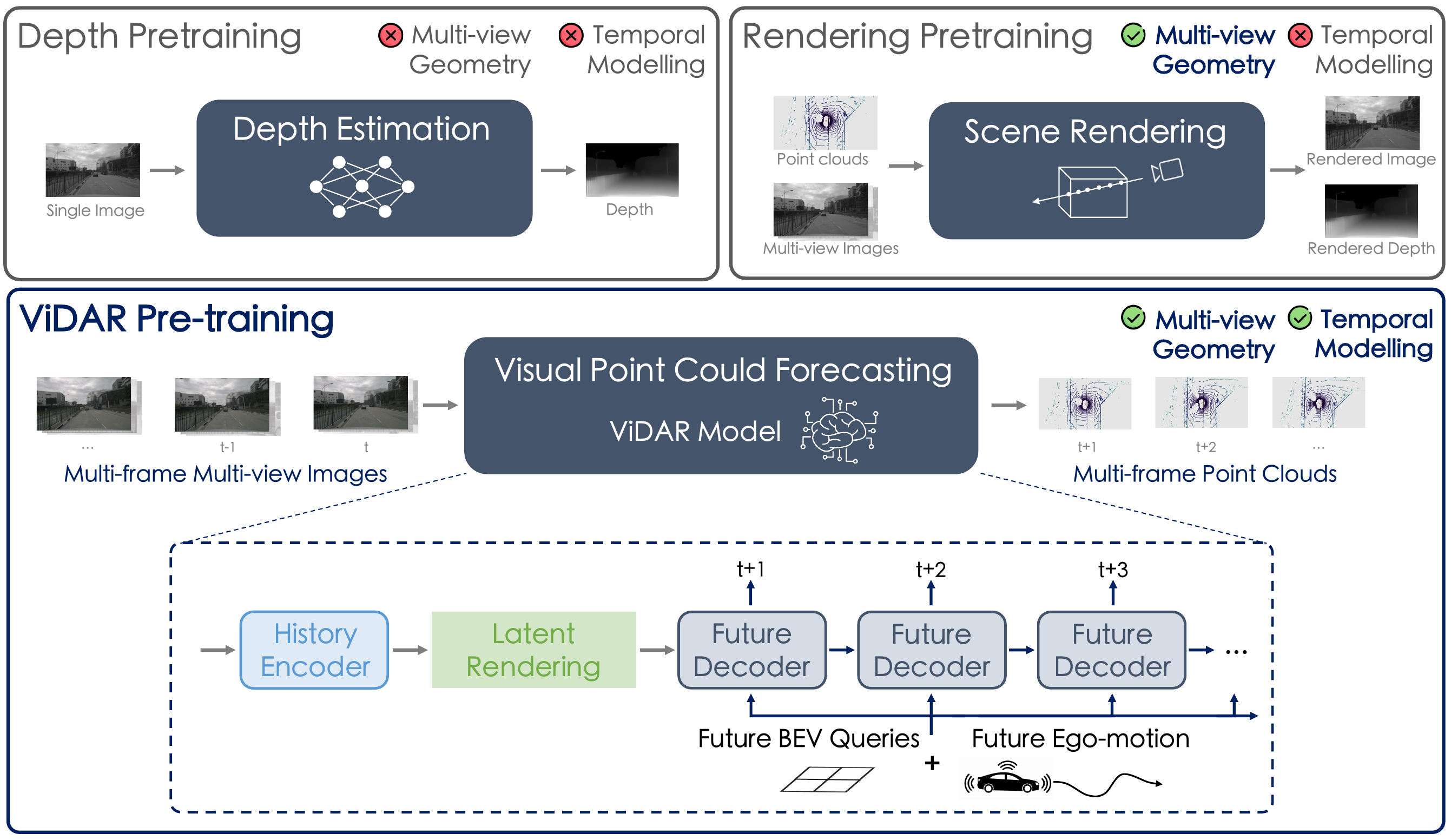}\\
  \vspace{-0.1in}
  \caption{\textbf{Comparisons among visual autonomous driving pre-training paradigms and our \algname architecture.} Compared to existing methods, visual point cloud forecasting jointly models multi-view geometry and temporal dynamics. We then propose \algname, using Image-LiDAR sequences to pre-train visual encoders.
  }
  \label{fig:teaser-pipeline}
  \vspace{-0.01in}
\end{figure}

Though extensive research of pre-training in computer vision has been conducted~\cite{he2020moco,chen2020mocov2,wu2018unsupervised,CMC,He2022MaskedAutoencoders,xie2020pointcontrast,hou2021exploring,jiang2023selfsupervised}, its application in visual autonomous driving remains seldom explored. Visual autonomous driving poses great challenges in pre-training as it requires the features to maintain semantics, 3D geometry, and temporal dynamics at the same time for joint perception, prediction, and planning~\cite{wu2023ppgeo,chen2023e2esurvey}.
As a result, most models still rely on supervised pre-training, like 3D detection \cite{wang2021fcos3d,park2021pseudolidar} or occupancy \cite{Tong_2023_ICCV, Min2023OccupancyMAE, Yan2023SPOT}, using labeled data that is often unavailable at scale~\cite{li2023bevsurvey}. Some approaches propose estimating depth \cite{park2021pseudolidar} or rendering masked scenes \cite{yang2023unipad} as pre-training. They use Image-LiDAR pairs as a means of achieving scalable annotation-free pre-training. However, they struggle in either multi-view 3D geometry or temporal modeling (\Cref{fig:teaser-pipeline}). Depth estimation retrieves depth from one image, limited in multi-view geometry; rendering techniques reconstruct scenes from multi-view images but lacking temporal modeling.
However, temporal modeling is crucial in end-to-end autonomous driving systems, \eg, UniAD \cite{hu2023_uniad}, especially for prediction and planning which are the ultimate goals and require accurate scene flow and object motion for decision-making. Due to the absence of temporal modeling, existing approaches are insufficient for pre-training the end-to-end system.

In this work, we explore pre-training tailored for end-to-end visual autonomous driving applications, including not only perception but also prediction and planning~\cite{chen2023e2esurvey}. We formulate a new pre-text task, visual point cloud forecasting (Figure~\ref{fig:teaser-pipeline}), to fully exploit information of semantics, 3D geometry, and temporal dynamics behind the raw Image-LiDAR sequences, with being scalable into consideration. It predicts future point clouds from historical visual images.

The main rationale of visual point cloud forecasting lies in the simultaneous supervision of semantics, 3D structure, and temporal modeling. By compelling the model to predict the future from history, it supervises the extraction of scene flow and object motion which are crucial for temporal modeling and future estimation. Meanwhile, it involves the reconstruction of point clouds from images, which supervises the multi-view geometry and semantic modeling. Therefore, features from visual point cloud forecasting embed information of both geometric and temporal hints, beneficial for perception, tracking, and planning simultaneously.

To this end, we present \textbf{\algname}, a general visual point cloud forecasting approach for pre-training (\Cref{fig:teaser-pipeline}). \algname includes three parts, \textit{\encoder}, \textit{\render} operator, and \textit{\decoder}.
The \encoder is the target structure for pre-training. It could be any visual BEV encoder \cite{li2022bevformer} to embed visual sequences into BEV space. These BEV features are sent to the \render operator. \render plays a crucial role in enabling \algname benefit downstream performance. It solves the ray-shaped BEV features issue \cite{Li2023FBBEV, Zeng_2023_CVPR}, models 3D geometric latent space, and bridges encoder and decoder. The \decoder is an auto-regressive transformer that takes historical BEV features to iteratively predict future point clouds for arbitrary timestamps.

\algname provides a comprehensive solution for visual autonomous driving pre-training. We test \algname on nuScenes dataset \cite{nuscenes2019} in terms of point cloud forecasting and downstream verifications. Though with visual inputs, \algname still outperforms previous forecasting methods using point clouds, $\sim$33\% Chamfer Distance reduction on point cloud estimation of 1s future. \algname also improves downstream performance. Using Image-LiDAR sequences only, \algname surprisingly outperforms 3D detection pre-training \cite{wang2021fcos3d}, \eg, by 1.1\% mAP and 2.5\% mIoU for detection and semantic occupancy prediction, under the same data scale. If also based on the 3D detection pre-training, \algname boosts previous methods by 4.3\% mAP and 4.6\% mIoU. Further, due to the effective pre-training on the joint capture of geometric and temporal information, \algname improves UniAD \cite{hu2023_uniad} on all tasks for end-to-end autonomous driving including perception, prediction, and planning by a large margin (\Cref{fig:teaser}). Experimental results validate that visual point cloud forecasting enables scalable autonomous driving.

\section{Related Work}
\label{sec:related-work}

\myparagraph{Pre-training for Visual Autonomous Driving.}
Pre-training for scalable applications has been extensively studied in general vision. These approaches can be roughly divided into contrastive approaches \cite{he2020moco,chen2020mocov2,CMC,khosla2020supervised}, which learn discriminative features from positive and negative pairs; and masked signal modeling \cite{He2022MaskedAutoencoders,devlin2018bert,wei2023masked,xie2021simmim}, which recover discarded signals from remained signals to capture a comprehensive understanding of global semantics.

In contrast, pre-training for visual autonomous driving is still under-explored. Visual autonomous driving poses great challenges as it requires semantic understanding, 3D structural awareness, and temporal modeling at the same time, for joint perception, prediction, and planning. Existing vision methods mainly consider semantics; methods based on Image-LiDAR pairs \cite{park2021pseudolidar,yang2023unipad} struggle with temporal modeling; other supervised strategies \cite{wang2021fcos3d,Tong_2023_ICCV} are not scalable.
In this work, we propose visual point cloud forecasting, which simultaneously models semantics, temporal dynamics, and 3D geometry by a uniform process, and is easily scaled up.

\myparagraph{Point Cloud Forecasting.}
Point cloud forecasting, one of the most fundamental self-supervised tasks for autonomous driving, predicts future point clouds from past point cloud inputs. Previous works use range image \cite{meyer2019lasernet,bewley2021range}, a representation obtained by projecting point clouds to dense 2D images using sensor intrinsic and extrinsic parameters. Based on historical range images, they apply 3D convolutions \cite{mersch2021corl}, or LSTMs \cite{weng2020inverting,weng2022s2net} to predict future point clouds. Yet, they additionally model the motion of sensor intrinsic and extrinsic parameters. Later methods factor out the estimation of sensors by introducing 4D occupancy prediction \cite{khurana2023point} and differentiable ray-casting \cite{khurana2022differentiable}, which ensures a better modeling of the world.
Compared to prior literature, we aim at visual point cloud forecasting, using past images to predict future point clouds. Meanwhile, we raise this task as a pre-training paradigm for visual autonomous driving and demonstrate its superiority in a wide range of downstream applications.

\section{Methodology}
\label{sec:method}
In this section, we elaborate on our \algname, a visual point cloud forecasting approach for general autonomous driving pre-training. We begin with an overview of \algname in \Cref{sec:sec3.1}, and subsequently delve into \render and \decoder in \Cref{sec:sec3.2} and \Cref{sec:sec3.3}, respectively.
 
\subsection{Overview}
\label{sec:sec3.1}
As depicted in \Cref{fig:teaser-pipeline}, \algname comprises three components: \textbf{(a)} a \textit{\encoder}, also the target structure of pre-training, which extracts BEV embeddings $\mathcal{F}_{\text{bev}}$ from visual sequence inputs $\mathcal{I}$; It can be any visual BEV encoder~\cite{li2022bevformer,huang2021bevdet,liu2022petr};
\textbf{(b)} a \textit{\render} operator, which simulates the volume rendering operation in latent space so as to obtain geometric embedding $\hat{\mathcal{F}}_{\text{bev}}$ from  $\mathcal{F}_{\text{bev}}$;
and \textbf{(c)} a \textit{\decoder}, which predicts future BEV features $\hat{\mathcal{F}}_{t}$ at timestamps $t\in\{1,2,...\}$ in an auto-regressive manner.
Finally, a prediction head is followed to project $\hat{\mathcal{F}}_{t}$ into 3D occupancy volume $\mathcal{P}_{t}$. This process is formulated as:
\begin{equation} \label{eq:overall}
	\begin{aligned}
		&\mathcal{F}_{\text{bev}} \ = \texttt{Encoder}(\mathcal{I}), \\
            &\hat{\mathcal{F}}_{\text{bev}} \ = \texttt{LatentRender}(\mathcal{F}_{\text{bev}}), \\
            &\hat{\mathcal{F}}_{t} \ \ \ \ = \texttt{Decoder}(\hat{\mathcal{F}}_{t-1}),\ \text{where}\   \hat{\mathcal{F}}_0 = \hat{\mathcal{F}}_{\text{bev}}, \\
            &\mathcal{P}_{t} \ \ \ \  = \texttt{Projection}(\hat{\mathcal{F}}_{t}).
	\end{aligned}
\end{equation}
Point cloud predictions are obtained from the predicted occupancy volume $\mathcal{P}_{t}$. This process is similar to the previous point cloud forecasting method \cite{khurana2023point}. Specifically, we first cast rays from the origin to various designated directions, then figure out the distance of waypoints along each ray with the maximum occupancy response, and finally compute the point position according to the distance and corresponding ray direction.

\subsection{\render}
\label{sec:sec3.2}
A straightforward solution of visual point cloud forecasting for pre-training is to incorporate the \encoder and \decoder directly with differentiable ray-casting~\cite{khurana2023point}, which is the crucial component in state-of-the-art point cloud forecasting methods to render point clouds from predicted occupancy volume and compute loss for backpropagation.
However, our experimental results show that this approach does not yield improvements and even has a detrimental effect on the downstream tasks due to the defective geometric feature modeling ability.

\paragraph{Preliminary.}
Differentiable ray-casting is a volume rendering process operating on an occupancy volume, denoted as $\mathcal{P} \in \mathbb{R}^{L \times H \times W}$. It renders depths of various rays and subsequently converts the depths with corresponding ray directions to point clouds.

Formally, starting from the origin sensor position, $\mathbf{o} \in \mathbb{R}^3$, differentiable ray-casting casts $n$ rays with varying directions, $\mathbf{d} \in \mathbb{R}^{n \times 3}$. Along each ray $i$, it uniformly samples $m$ waypoints at different distances $\lambda^{(j)} \in \mathbb{R}, j\in \{1, 2, ..., m\}$ until reaching the boundary of the 3D space. The coordinates of these waypoints are calculated as:
\begin{equation} \label{eq:diff_raycasting_coord}
	\begin{aligned}
		\mathbf{x}^{(i, j)} = \mathbf{o} + \lambda^{(j)} \mathbf{d}^{(i)}.
	\end{aligned}
\end{equation}
Those waypoint coordinates, $\mathbf{x} \in \mathbb{R}^{n \times m \times 3}$, are used to compute occupancy values. This process is quantized  \cite{khurana2023point}, wherein the waypoints are discretized to occupancy volume grids.
Then, the values of waypoints are derived as the associated values of volume grids, $\mathbf{p}^{(i, j)} = \mathcal{P}^{([\mathbf{x}^{(i,j)}])}$. Here
$[\cdot]$ denotes a rounding operation for discretizing waypoints.

Differentiable ray-casting renders the corresponding depth of the $i$-th ray, $\hat{\mathbf{\lambda}}^{(i)}$, by an integral process:
\begin{equation} \label{eq:diff_raycasting_prob}
	\begin{aligned}
		\hat{\mathbf{p}}^{(i, j)} = 
                % \left (
                \bigg [
                \prod_{k=1}^{j-1}(1 - \mathbf{p}^{(i, k)})
                %\right ) 
                \bigg ]
                \mathbf{p}^{(i, j)},
	\end{aligned}
\end{equation}
\vspace{-.1in}
\begin{equation} \label{eq:diff_raycasting_depth}
	\begin{aligned}
		\hat{\mathbf{\lambda}}^{(i)} =
                \sum_{j=1}^{m}\hat{\mathbf{p}}^{(i,j)} \lambda^{(j)}.
	\end{aligned}
\end{equation}

For simplicity, we name Eq.~\ref{eq:diff_raycasting_prob} and Eq.~\ref{eq:diff_raycasting_depth} as \textit{conditional probability function} and \textit{distance expectation function}. The conditional probability function determines the occupancy of a grid by considering the conditional probability that prior waypoints are unoccupied and the ray terminates at this particular grid; the distance expectation function retrieves depths from the occupancy of grids in 3D volume. L1 loss is then applied to supervise the rendered depth for training point cloud forecasting.

\begin{figure}[t]
  \centering
  % \vspace{-0.1in}
  \includegraphics[width=0.95\linewidth]{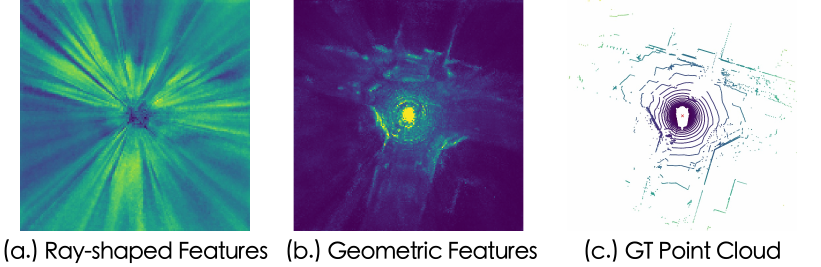}\\
  \vspace{-0.1in}
  \caption{\textbf{Ray-shaped Features \vs Geometric Features.} Ray-shaped features show similar feature responses on BEV grids along the same ray;
  while geometric features from the \render maintain discriminative 3D geometry and can describe the 3D world in latent space.
  }
  \label{fig:method_latentrender}
  % \vspace{-0.1in}
\end{figure}

Despite the great success of differentiable ray-casting in the task of point cloud forecasting, its application in visual point cloud forecasting pre-training does not bring any benefit for downstream performance (Table~\ref{tab:diff_raycasting}). After such pre-training, ray-shaped features~\cite{Li2023FBBEV,Zeng_2023_CVPR}, where grids along the same ray tend to possess similar features (\Cref{fig:method_latentrender} - (a.)), are observed. The underlying reason is that waypoints along the same ray in 3D space usually correspond to the same pixel in the visual image, resulting in a tendency to learn similar feature responses. As a consequence,
these ray-shaped features are not discriminative and representative enough when transferred to downstream applications, leading to reduced performance.

\begin{figure}[t]
  \centering
  % \vspace{-0.1in}
  \includegraphics[width=1.0\linewidth]{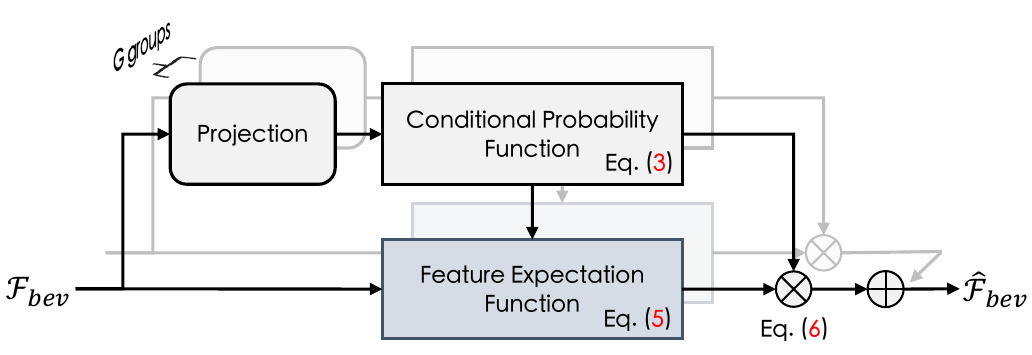}\\
  \vspace{-0.1in}
  \caption{\textbf{Multi-group \render} comprises several \render running in parallel for different channels. \render captures geometric features by the conditional probability function and the feature extraction function. ``$\bigoplus$'' means concatenating multi-group features among channel dimensions.
  }
  \label{fig:method_lr_structure}
  \vspace{-0.1in}
\end{figure}

\myparagraph{\render.}
\begin{table}[t]
  \centering
  \setlength{\tabcolsep}{4.5mm}
  \scalebox{0.8}{
  \begin{tabular}{l|ccc}
    \toprule[1.5pt]
    \makecell[l]{Forecasting \\ Structure} & 
    N/A & \makecell[l]{Differentiable \\ Ray-casting} &  
    \makecell[l]{Latent \\ Rendering}\\
     \midrule
     NDS (\%) & 44.11 & 40.20 ({\bf -3.91} ) & 47.58 ({\bf +3.47} ) \\
    \bottomrule[1.5pt]
  \end{tabular}
  }
  \vspace{-.1in}
  \caption{\textbf{Downstream detection performance under different forecasting structures.} ``N/A'' represents the baseline without forecasting pre-training.
  We observe a performance drop when directly using \encoder and \decoder with differentiable ray-casting for pre-training; while, with the \render operator, the performance is significantly improved.
  }
  \label{tab:diff_raycasting}
\end{table}
In order to extract more discriminative and representative features, we introduce the \render operator. It first computes the ray-wise feature through a \textit{feature expectation function}, then customizes features of each grid by weighting the ray-wise feature with its associated conditional probability. The overall structure is depicted in \Cref{fig:method_lr_structure}.

To be specific, inspired by Eq.~\eqref{eq:diff_raycasting_depth}, the feature expectation function is formulated in a similar form:
\begin{equation} \label{eq:latentrender_ray}
	\begin{aligned}
		\hat{\mathcal{F}}^{(i)} &=
                \sum_{k=1}^{m}\hat{\mathbf{p}}^{(i, k)} \mathcal{F}_{\text{bev}}^{(k)},
	\end{aligned}
\end{equation}
where $i$ represents the ray extending from the origin to the $i$-th BEV grid. Here, $\hat{\mathbf{p}}$ is the conditional probability, computed through the conditional probability function Eq. \eqref{eq:diff_raycasting_prob}, which takes as input the learnable independent probability projected from $\mathcal{F}_{bev}$. The ray-wise features are shared by all grids lying in the same ray.

Then, we compute the grid feature as:
\begin{equation} \label{eq:latentrender_mul}
	\begin{aligned}
            \hat{\mathcal{F}}_{\text{bev}} = \hat{\mathbf{p}} \cdot \hat{\mathcal{F}},
	\end{aligned}
\end{equation}
which highlights the response of BEV grids with higher conditional probability so as to make $\hat{\mathcal{F}}_{bev}$ discriminative. This enables the BEV encoder to learn the geometric features during pre-training (\Cref{fig:method_latentrender} - (b.)).

To enhance the diversity of geometric features, we further design the multi-group \render. By parallelizing multiple \render on different feature channels, we allow ray-wise features to maintain diverse information, leading to better downstream performance.

As described in Eq. \eqref{eq:diff_raycasting_prob}, the conditional probability of each BEV grid is determined not only by its own independent response, but also by the response of all its prior grids. Consequently, in the pre-training phase, once the model raises the response of a particular BEV grid, the corresponding responses of all its prior and subsequent grids are suppressed, which mitigates the issue of ray-shaped features during pre-training. After the pre-training with Latent Rendering, it is generally observed that there are only a few peaks with higher responses on a specific ray, indicating the presence of objects or structures in the scene. This effectively promotes a more accurate and consistent understanding of the 3D environment.

\begin{figure}[t]
  \centering
  \includegraphics[width=1.0\linewidth]{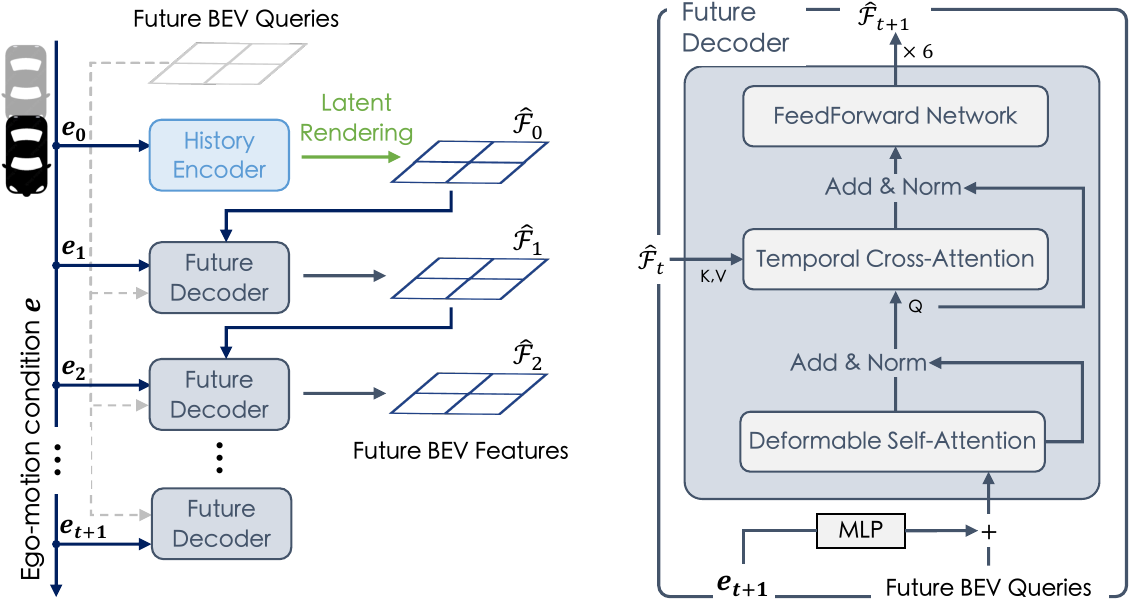}\\
  \vspace{-0.1in}
  \caption{\textbf{\decoder} iteratively predicts the next BEV features, $\hat{\mathcal{F}}_t$, from the conditions of ego-motion $\mathbf{e}_{t}$ and the last BEV features, to enable specific future predictions with any ego-control.
  }
  \label{fig:method_decoder}
  \vspace{-0.1in}
\end{figure}
\subsection{\decoder}
\label{sec:sec3.3}
The \decoder predicts the next BEV features $\hat{\mathcal{F}}_{t}$ of frame $t$ based on the inputs of previous BEV latent space $\hat{\mathcal{F}}_{t-1}$ and the expected next ego-motion, $\mathbf{e}_{t}$. The predicted features are then used to generate point clouds as Eq. \eqref{eq:overall}.

\myparagraph{Architecture.}  
As depicted in \Cref{fig:method_decoder}, \decoder is a transformer that can be iteratively used to predict future BEV features from the last roll-out embeddings in an auto-regressive manner. In the $t$-th iteration, it first encodes the ego-motion condition $\mathbf{e}_{t}$, which describes the expected coordinates and heading of ego-vehicle in the next frame, into high-dimensional embeddings by multi-layer perceptron (MLP), which are then added to future BEV queries as inputs of the transformer. Then, $6$ transformer layers, composed of a Deformable Self-Attention \cite{zhu2020deformable}, a Temporal Cross-Attention and a FeedForward Network \cite{VIT}, are used to predict the future $\hat{\mathcal{F}}_{t}$ based on the condition and the last BEV features $\hat{\mathcal{F}}_{t-1}$.

The Temporal Cross-Attention layer follows the design of Deformable Cross-Attention \cite{zhu2020deformable}. The difference lies in the reference coordinates of query points. In the context of Deformable Cross-Attention \cite{zhu2020deformable}, ``reference coordinates'' refer to the corresponding positions of query points on the feature maps of keys and values. Typically, they are consistent. However, as for \decoder, due to the moving of the ego-vehicle, the ego-coordinate systems between the last and target frame are not necessarily aligned. Therefore, we additionally compute the reference coordinates of future BEV queries in previous BEV feature maps, according to the ego-motion condition, to align coordinate systems.

After obtaining the next BEV features $\hat{\mathcal{F}}_{t}$, we use a projection layer to generate the occupancy volume $\mathcal{P}_t$. 

\myparagraph{Loss.}
Instead of using L1 loss to supervise the depths of various rays, we directly apply ray-wise cross-entropy loss to maximize the response of points along its corresponding ray, as we have already obtained geometric features after the \render operator.
To be specific, for each ground-truth point of the $t$-th future point clouds, we cast a ray from the origin position $\mathbf{o}$ (the sensor position) towards the point, uniformly sample some waypoints along the ray until out of the volume, and compute cross-entropy loss for the ray to maximize the response of the point position and minimize the response of other waypoint positions. This process is formulated as:
\begin{equation} \label{eq:loss}
	\begin{aligned}
            \mathcal{L} = -
                \frac{1}{Tn}
                \sum_{t=1}^{T}
                \sum_{i=1}^{n} \text{log}\left(
                    \frac{e^{\mathcal{P}_t^{(\mathbf{g}^{(i)})}}}{\sum_je^{\mathcal{P}_t^{(\mathbf{x}^{(i,j)})}}
                        + e^{\mathcal{P}_t^{(\mathbf{g}^{(i)})}}}
                \right),
	\end{aligned}
\end{equation}
where $T$, $n$ indicate the number of future supervisions and the number of points in the $t$-th ground-truth point clouds. $\mathbf{g}^{(i)}$ and $\mathbf{x}^{(i,j)}$ are the coordinates of $i$-th ground-truth point and $j$-th waypoints along the same ray. $\mathcal{P}_t^{(\cdot)}$ is trilinear interpolation to obtain corresponding values from volume $\mathcal{P}_t$.

\section{Experiments}
This section investigates the following questions:

\begin{itemize}\vspace{-0.5pt}

\item Can future point clouds be estimated from visual history, and how about \algname compared to point cloud methods?

\vspace{0.05in}
\item Can \algname help perception, prediction, and planning at the same time so as for scalable autonomous driving?

\vspace{0.05in}
\item Can \algname reduce the reliance of downstream applications on precise human annotations?

\vspace{0.05in}
\item How do different modules affect final performance?
\end{itemize}

\subsection{Setup}

\myparagraph{Dataset.}
We conduct experiments on the challenging nuScenes \cite{nuscenes2019} dataset, which is a large-scale dataset with 1,000 autonomous driving sequences. This dataset is widely utilized in perception tasks including 3D object detection \cite{li2022bevformer,huang2021bevdet,li2023bevdepth,Li_2023_ICCV}, multi-object tracking \cite{Hu2021QD3DT,zhang2022mutr3d,pang2021simpletrack}, and semantic occupancy prediction \cite{Tong_2023_ICCV,openscene2023}. It has also become a popular benchmark for subsequent research on end-to-end autonomous driving, including map segmentation \cite{wang2023openlanev2,li2023toponet}, trajectory predictions \cite{vip3d,pnpnet}, future occupancy prediction \cite{fiery2021,zhang2022beverse}, and open-loop planning \cite{hu2022stp3,hu2023_uniad,jiang2023vad}.

\myparagraph{Implementation Details.}
We base our implementation on mmDet3D codebase \cite{mmdet3d2020} and conduct downstream verifications on BEVFormer for 3D detection, OccNet \cite{Tong_2023_ICCV} for semantic occupancy prediction, and UniAD \cite{hu2023_uniad} for unified perception, prediction, and planning. We choose those downstream baselines due to their effectiveness on a wide range of tasks and sharing the same BEV encoder structure, BEVFormer encoder \cite{li2022bevformer}. Without any specifications, the default historical encoder of \algname is the BEVFormer-base encoder, consisting of a ResNet101-DCN \cite{ResNet,dai2017deform} backbone with an FPN neck \cite{FPN} and additional 6 encoder layers to extract BEV features from multi-view image sequences, which is consistent with downstream models.

To render geometric features, we use a 16-group \render. Each group is responsible for rendering latent spaces of 16 channels given the features of 256 channels after the BEVFormer-base encoder.
The \decoder is a 6-layer structure with a channel of 256 for each. The future BEV queries are $200 \times 200$ learnable tokens indicating a valid perception range of [-51.2m, 51.2m] for the $X$ and $Y$ axis. We then use a projection layer with output channels as $16$ to transform the predicted future BEV features to occupancy volume prediction $\mathcal{P} \in \mathbb{R}^{200 \times 200 \times 16}$, where $16$ indicates the height dimension with a range of [-5m, 3m].

During pre-training, we use 5 frames of historical multi-view images and iterate the \decoder 6 times to predict point clouds for the future 3 seconds (each frame has 0.5 second interval). In each training step, we randomly select 1 future prediction for computing loss and detach the gradients of the others to save GPU memory. We pre-train the system for 50 epochs by AdamW optimizer \cite{loshchilov2019decoupled,adam} with an initial learning of 2e-4 adjusted by cosine annealing strategy. For fine-tuning, we follow the same training strategy as the officially released downstream models.

\subsection{Main Results}
We now demonstrate the effectiveness of \algname across different tasks. First, we test the ability of \algname as a point cloud forecasting framework and compare it with the state-of-the-art approach that uses LiDAR inputs. Then, we show its advancement as a visual autonomous driving pre-training solution. We report the downstream comparison results in the order of perception-prediction-planning with previous state-of-the-art models on the nuScenes validation dataset.

\myparagraph{Downstream Settings.}
For downstream verifications, we test \algname to pre-train BEV encoders under different initialization settings, listed as the following:

\begin{itemize}\vspace{-0.5pt}

\item \textbf{\algname-cls:} The BEV encoders are initialized with the backbone pre-trained for ImageNet classification~\cite{imagenet_cvpr09}, followed by \algname pre-training on nuScenes dataset.

\vspace{0.05in}
\item \textbf{\algname-2D-det:} The BEV encoder backbones are first pre-trained for 2D object detection on COCO dataset~\cite{lin2014coco} before \algname pre-training on nuScenes dataset.

\vspace{0.05in}
\item \textbf{\algname-3D-det:} The BEV encoder backbones are pre-trained first by 3D detection on the nuScenes dataset, using FCOS3D~\cite{wang2021fcos3d}, before \algname pre-training.
\end{itemize}
For UniAD experiments, after \algname pre-training, we first fine-tune BEVFormer for 3D detection, which is then used as the initialization for subsequent two-stage fine-tuning, consistent with the UniAD official implementation.

\begin{table}[t]
    \centering
    % \vspace{-0.1in}
    \setlength{\tabcolsep}{1.2mm}
    \scalebox{0.8}{
    \begin{tabular}{c|l|c|cccccc}
        \toprule[1.5pt]
        \multicolumn{1}{c|}{ \multirow{2}{*}{\makecell[c]{History \\ Horizon}}} & 
        \multicolumn{1}{c|}{ \multirow{2}{*}{Method}} & 
        \multicolumn{1}{c|}{ \multirow{2}{*}{Modality}} &
            \multicolumn{6}{c}{ \multirow{1}{*}{Chamfer Distance (m$^2$) $\downarrow$}} \\
        & & &  0.5s & 1.0s & 1.5s & 2.0s & 2.5s & 3.0s \\
        \midrule
        \multicolumn{1}{c|}{ \multirow{2}{*}{1s}} & 4D-Occ \cite{khurana2023point} & L & 1.26 & 1.88 & - & - & - & -  \\
        & \algname & \cellcolor{lightgray}C & \cellcolor{lightgray}\bf 1.11 & \cellcolor{lightgray}\bf 1.25 & \cellcolor{lightgray}\bf 1.40 & \cellcolor{lightgray}\bf 1.57 & \cellcolor{lightgray}\bf 1.76 & \cellcolor{lightgray}\bf 1.97 \\
        \midrule
        \multicolumn{1}{c|}{ \multirow{2}{*}{3s}} & 4D-Occ \cite{khurana2023point} & L & \bf0.91 & 1.13 & 1.30 & 1.53 & 1.72 & 2.11  \\
        & \algname & \cellcolor{lightgray}C & \cellcolor{lightgray} 1.01 & \cellcolor{lightgray}\bf1.12 & \cellcolor{lightgray}\bf1.25 & \cellcolor{lightgray}\bf1.38 & \cellcolor{lightgray}\bf1.54 & \cellcolor{lightgray}\bf1.73 \\
        \bottomrule[1.5pt]
    \end{tabular}
    }
    \vspace{-0.1in}
    \caption{\textbf{Point cloud forecasting.} \algname surpasses prior state-of-the-art method on future prediction, using visual input only.}
    % \vspace{-.1in}
    \label{tab:pc_forecast}
\end{table}
\paragraph{Point Cloud Forecasting.}
In \Cref{tab:pc_forecast}, we present comparisons between our \algname and the previous state-of-the-art point cloud forecasting method, 4D-Occ \cite{khurana2023point}. The evaluation metric is Chamfer Distance. We evaluate both methods using input time horizons of 1s and 3s, corresponding to input sequences of 2 frames and 6 frames, respectively, following the same setup as in 4D-Occ. To provide detailed comparisons of performance, we report the quantitative forecasting results for each future timestamp. Only points within the range of [-51.2m, 51.2m] on the X- and Y-axis are considered during evaluation.

As presented in \Cref{tab:pc_forecast}, \algname consistently outperforms 4D-Occ on both 1s and 3s settings, despite utilizing visual inputs exclusively. Specifically, with 1s history input, \algname achieves remarkable improvement over 4D-Occ, reducing forecasting errors by $\sim$33\% for future 1s predictions. When using 3s inputs, we observe a $\sim18\%$ error reduction for 3s forecasting. Moreover, due to the auto-regressive design of our \decoder, \algname effectively predicts arbitrary future, though with the constraint of a limited 1s input horizon. These experiments demonstrate the effectiveness of \algname for point cloud forecasting.

\begin{table}[t]
  \centering
  %\vspace{-.1in}
  \setlength{\tabcolsep}{1.2mm}
  \scalebox{0.8}{
  \begin{tabular}{{l|l|l|cc}}
    \toprule[1.5pt]
    Methods  & \multicolumn{1}{c|}{ \multirow{1}{*}{Encoder}} & \multicolumn{1}{c|}{ \multirow{1}{*}{Pre-train}} & mAP (\%) & NDS (\%) \\
    \midrule
    \multicolumn{1}{c|}{ \multirow{11}{*}{\makecell[l]{BEV-\\Former} \cite{li2022bevformer}}} & \multicolumn{1}{c|}{ \multirow{2}{*}{RN50 \cite{ResNet}}} & ImageNet-cls \cite{imagenet_cvpr09} & 25.2 & 35.4 \\
     & & \algname-cls & \cellcolor{lightgray}\bf 29.0 & \cellcolor{lightgray}\bf 38.8 \\
     \cmidrule{2-5}
     & \multicolumn{1}{c|}{ \multirow{4}{*}{RN101 \cite{ResNet}}} & ImageNet-cls \cite{imagenet_cvpr09} & 37.7 & 47.7  \\
     &  & \algname-cls & \cellcolor{lightgray}\bf 42.6 & \cellcolor{lightgray}\bf 51.8 \\
     \cmidrule{3-5}
     &  & nus-3D-det \cite{wang2021fcos3d} & 41.5 & 51.7 \\ 
     &  & \algname-3D-det & \cellcolor{lightgray}\bf 45.8 & \cellcolor{lightgray}\bf 54.8 \\
     \cmidrule{2-5}
     & \multicolumn{1}{c|}{ \multirow{2}{*}{Intern-S \cite{wang2022internimage}}} & COCO-2D-det \cite{lin2014coco} & 41.5 & 51.2 \\
     & & \algname-2D-det & \cellcolor{lightgray}\bf 47.6  & \cellcolor{lightgray}\bf 56.4 \\
     \cmidrule{2-5}
     & \multicolumn{1}{c|}{ \multirow{2}{*}{Intern-B \cite{wang2022internimage}}} & COCO-2D-det \cite{lin2014coco} & 42.9 & 52.0 \\
     & & \algname-2D-det & \cellcolor{lightgray}\bf 50.1 & \cellcolor{lightgray}\bf 57.6 \\
    \bottomrule[1.5pt]
  \end{tabular}
  }
  \vspace{-0.1in}
  \caption{\textbf{
  Detection performance} of BEVFormer with and without \algname pre-training under various backbones and initializations.
  }
  \label{tab:detection}
\end{table}

\paragraph{Perception.}
We verify \algname on four downstream perception tasks, 3D object detection, semantic occupancy prediction, map segmentation, and multi-object tracking. In \Cref{tab:detection} and \Cref{tab:ssc}, we compare the performance of BEVFormer \cite{li2022bevformer} and OccNet \cite{Tong_2023_ICCV} with and without \algname pre-training under different backbones and initialization settings for 3D detection and semantic occupancy prediction, respectively. Notably, \algname, using solely Image-LiDAR sequences, outperforms 3D detection supervised pre-training (The 4th \& the 5th row in \Cref{tab:detection}, 42.6\% mAP to 41.5\% mAP; The 2nd \& the 3rd row in \Cref{tab:ssc}, 29.57\% mIoU to 26.98\% mIoU). Furthermore, we also observe huge improvements in map segmentation (\Cref{tab:map_seg}) and multi-object tracking (\Cref{tab:track}). These experiments demonstrate the effectiveness of \algname as a scalable pre-training method for enhancing 3D geometry modeling.

\begin{table}[t]
  \centering
  %\vspace{-0.1in}
  \setlength{\tabcolsep}{2.5mm}
  \scalebox{0.8}{
  \begin{tabular}{l|l|l|c}
    \toprule[1.5pt]
    Methods  & \multicolumn{1}{c|}{ \multirow{1}{*}{Encoder}} &  \multicolumn{1}{c|}{ \multirow{1}{*}{Pre-train}} & mIoU (\%) $\uparrow$ \\
    \midrule
    \multicolumn{1}{c|}{ \multirow{9}{*}{OccNet \cite{Tong_2023_ICCV}}} & \multicolumn{1}{c|}{ \multirow{4}{*}{RN101 \cite{ResNet}}} & ImageNet-cls \cite{imagenet_cvpr09} & 24.35 \\
     & & \algname-cls &  \cellcolor{lightgray}\bf 29.57 \\
    \cmidrule{3-4}
    &  & nus-3D-det \cite{wang2021fcos3d} & 26.98 \\
     &  & \algname-3D-det & \cellcolor{lightgray}\bf 31.67 \\
     \cmidrule{2-4}
     & \multicolumn{1}{c|}{ \multirow{2}{*}{Intern-S \cite{wang2022internimage}}} & COCO-2D-det \cite{lin2014coco} & 24.92 \\
     & & \algname-2D-det & \cellcolor{lightgray}\bf 30.51  \\
     \cmidrule{2-4}
     & \multicolumn{1}{c|}{ \multirow{2}{*}{Intern-B \cite{wang2022internimage}}} & COCO-2D-det \cite{lin2014coco} & 25.24  \\
     & & \algname-2D-det & \cellcolor{lightgray}\bf 31.69 \\
    \bottomrule[1.5pt]
  \end{tabular}
  }
  \vspace{-0.1in}
  \caption{\textbf{Semantic Occupancy Prediction.} \algname consistently improves OccNet on different backbones and initializations.}
  \label{tab:ssc}
\end{table}
\begin{table}[t]
  \centering
  \vspace{-.1in}
  \setlength{\tabcolsep}{3.5mm}
  \scalebox{0.8}{
  \begin{tabular}{l|c|c|c}
    \toprule[1.5pt]
    Methods  & Encoder & Pre-train & Lanes (\%) $\uparrow$ \\
    \midrule
    BEVFormer \cite{li2022bevformer} & RN101 & nus-3D-det \cite{wang2021fcos3d} & 23.9 \\
    \midrule
    \multicolumn{1}{l|}{ \multirow{2}{*}{UniAD \cite{hu2023_uniad}}} & 
    \multicolumn{1}{c|}{ \multirow{2}{*}{RN101}}
    & nus-3D-det \cite{li2022bevformer} & 31.3 \\
    & & \multicolumn{1}{l|}{ \multirow{1}{*}{\algname-3D-det}}  & \cellcolor{lightgray}\bf 33.2
    \\
    \bottomrule[1.5pt]
  \end{tabular}
  }
  \vspace{-0.1in}
  \caption{\textbf{Map segmentation.} \algname improves UniAD on online mapping. The metric is segmentation IoU.
  }
  \label{tab:map_seg}
\end{table}
\begin{table}[t]
  \centering
  \vspace{-.1in}
  \setlength{\tabcolsep}{1.7mm}
  \scalebox{0.8}{
  \begin{tabular}{l|l|l|c}
    \toprule[1.5pt]
    Methods  & \multicolumn{1}{c|}{ \multirow{1}{*}{Encoder}} & \multicolumn{1}{c|}{ \multirow{1}{*}{Pre-train}} & AMOTA (\%) $\uparrow$ \\
    \midrule
    ViP3D \cite{vip3d} & RN50 & nus-3D-det \cite{detr3d} & 21.7 \\
    QD3DT \cite{Hu2021QD3DT} & RN101 & - & 24.2 \\
    MUTR3D \cite{zhang2022mutr3d} & RN101 & ImageNet-cls & 29.4 \\
    DQTrack-DETR3D \cite{Li_2023_ICCV} & RN101 & nus-3D-det \cite{detr3d} & 36.7 \\
    DQTrack-UVTR \cite{Li_2023_ICCV} & RN101 & nus-3D-det \cite{li2022uvtr} & 39.6 \\
    DQTrack-Stereo \cite{Li_2023_ICCV} & RN101 & nus-3D-det \cite{li2023bevstereo} & 40.7 \\
    DQTrack-PETRv2 \cite{Li_2023_ICCV} & V2-99 & nus-3D-det \cite{liu2022petrv2} & 44.6 \\
    \midrule
    \multicolumn{1}{l|}{ \multirow{2}{*}{UniAD-Stage1 \cite{hu2023_uniad}}} & 
    \multicolumn{1}{l|}{ \multirow{2}{*}{RN101}}
    & nus-3D-det \cite{li2022bevformer} & 39.0 \\
    & & \algname-3D-det  & \cellcolor{lightgray}\bf 45.1
    \\
    \bottomrule[1.5pt]
  \end{tabular}
  }
  \vspace{-.1in}
  \caption{\textbf{Multi-object tracking.} With \algname, UniAD outperforms previous end-to-end trackers using solely visual images.}
  \label{tab:track}
\end{table}

\begin{comment}
\begin{table}[t]
  \centering
  \setlength{\tabcolsep}{0.7mm}
  \scalebox{0.8}{
  \begin{tabular}{l|l|cccc}
    \toprule[1.5pt]
    Methods  & \multicolumn{1}{c|}{ \multirow{1}{*}{Encoder}} & AMOTA $\uparrow$ & AMOTP $\downarrow$ & IDS $\downarrow$ \\
    \midrule
    ViP3D \cite{vip3d} & RN50 & 0.217 & 1.625 & - \\
    QD3DT \cite{Hu2021QD3DT} & RN101 & 0.242 & 1.518 & - \\
    MUTR3D \cite{zhang2022mutr3d} & RN101 & 0.294 & 1.498 & 3822 \\
    DQTrack-DETR3D \cite{Li_2023_ICCV} & RN101 & 0.367 & 1.351 & 1120 \\
    DQTrack-UVTR \cite{Li_2023_ICCV} & RN101 & 0.396 & 1.310 & 1267 \\
    DQTrack-Stereo \cite{Li_2023_ICCV} & RN101 & 0.407 & 1.318 & 1003 \\
    DQTrack-PETRv2 \cite{Li_2023_ICCV} & V2-99 &  0.446 & 1.251 & 1193 \\
    \midrule
    UniAD-Stage1 & RN101 & 0.390 & - & - \\
    \algname & RN101 & \cellcolor{lightgray}\bf 0.451 & \cellcolor{lightgray}\bf 1.232 & \cellcolor{lightgray}\bf 1031
    \\
    \bottomrule[1.5pt]
  \end{tabular}
  }
  \caption{Comparison of different 3D detection models on the nuScenes \textit{val} set.}
  \label{tab:track}
\end{table}
\end{comment}

\paragraph{Prediction.}
The motion forecasting comparisons are presented in \Cref{tab:motion_traj}. As shown, \algname significantly improves the performance of UniAD \cite{hu2023_uniad}. For instance, we observe a $\sim$10\% error reduction in minADE and a 3.5\% improvement in EPA.
In \Cref{tab:motion_occ}, we provide the comparison between UniAD with and without \algname pre-training in future occupancy prediction. The results demonstrate that \algname enhances the performance of UniAD for all areas. We observe improvements of 2.4\% IoU and 2.7\% VPQ for nearby areas, as well as improvements of 2.0\%  IoU and 2.5\% VPQ for distant areas. With \algname pre-training, UniAD overcomes its limitation in occupancy forecasting for distant objects and now outperforms BEVerse \cite{zhang2022beverse} in all areas. 
These experiments highlight the effectiveness of \algname in enhancing downstream models in utilizing temporal information and improving their prediction performance.

\begin{table}[t]
  \centering
  % \vspace{-.1in}
  \setlength{\tabcolsep}{2.8mm}
  \scalebox{0.8}{
  \begin{tabular}{l|cccc}
    \toprule[1.5pt]
    Methods & minADE (m) $\downarrow$ & minFDE (m) $\downarrow$ & MR $\downarrow$ & EPA $\uparrow$ \\
    \midrule
    PnPNet \cite{pnpnet} & 1.15 & 1.95 & 0.226 & 0.222 \\
    ViP3D \cite{vip3d} & 2.05 & 2.84 & 0.246 & 0.226 \\
    \midrule
    UniAD \cite{hu2023_uniad} & 0.75 & 1.08 & 0.158 & 0.463 \\
    \algname & \cellcolor{lightgray}\textbf{0.67} & \cellcolor{lightgray}\textbf{0.99} & \cellcolor{lightgray}\textbf{0.149} & \cellcolor{lightgray}\textbf{0.498}
    \\
    \bottomrule[1.5pt]
  \end{tabular}
  }
  \vspace{-.1in}
  \caption{\textbf{Motion forecasting}. \algname effectively enhances temporal modeling, which in turn boosts UniAD in future motion forecasting, showing consistent improvements on different metrics.}
  \label{tab:motion_traj}
\end{table}
\begin{table}[t]
  \centering
  \vspace{-0.1in}
  \setlength{\tabcolsep}{2.8mm}
  \scalebox{0.8}{
  \begin{tabular}{l|cccc}
    \toprule[1.5pt]
    Methods & VPQ-n. $\uparrow$ & VPQ-f. $\uparrow$ & IoU-n. $\uparrow$ & IoU-f. $\uparrow$ \\
    \midrule
    Fiery \cite{fiery2021} & 50.2 & 29.9 & 59.4 & 36.7 \\
    StretchBEV \cite{Akan2022ECCV} & 46.0 & 29.0 & 55.5 & 37.1 \\
    ST-P3 \cite{hu2022stp3} & - & 32.1 & - & 38.9 \\
    BEVerse \cite{zhang2022beverse} & 54.3 & 36.1 & 61.4 & 40.9 \\
    \midrule
    UniAD \cite{hu2023_uniad} & 54.6 & 33.9 & 62.8 & 40.1 \\
    \algname & \cellcolor{lightgray}\textbf{57.3} & \cellcolor{lightgray}\textbf{36.4} & \cellcolor{lightgray}\textbf{65.4} & \cellcolor{lightgray}\textbf{42.1} 
    \\
    \bottomrule[1.5pt]
  \end{tabular}
  }
  \vspace{-.1in}
  \caption{\textbf{Future occupancy prediction.} \algname improves UniAD on future occupancy prediction for objects in both near (noted as ``n.'', 30x30m) and far (noted as ``f.'', 50x50m) evaluation areas.}
  \label{tab:motion_occ}
\end{table}
\begin{table}[t]
  \centering
  \vspace{-.1in}
  \setlength{\tabcolsep}{2.8mm}
  \scalebox{0.8}{
  \begin{tabular}{l|c|c|c}
    \toprule[1.5pt]
    Methods & Modality & Avg.Col. (3s) (\%) $\downarrow$ & Avg.L2 (3s) (m) $\downarrow$ \\
    \midrule
    FF \cite{hu2021safe} & L & 0.43 & 1.43 \\
    EO \cite{khurana2022differentiable} & L & 0.33 & 1.60 \\
    ST-P3 \cite{hu2022stp3} & C & 0.71 & 2.11 \\
    VAD \cite{jiang2023vad} & C & 0.41 & 1.05 \\
    \midrule
    UniAD \cite{hu2023_uniad} & C & 0.27 & 1.12 \\
    \algname & C & \cellcolor{lightgray}\bf 0.23 & \cellcolor{lightgray}\bf 0.91 \\
    \bottomrule[1.5pt]
  \end{tabular}
  }
  \vspace{-.1in}
  \caption{\textbf{Planning.} \algname improves UniAD in terms of both collision avoidance and planning accuracy. Note that, the reported numbers are obtained by averaging the results of each timestamp in the future 3 seconds, which is consistent with the reported results of VAD \cite{jiang2023vad} and ST-P3 \cite{hu2022stp3} at the 3s timestamp instead of the averaged one. Please refer to \href{https://github.com/OpenDriveLab/UniAD/issues/29}{GitHub:Issue} for more details.
  }
  \label{tab:planning}
\end{table}
\paragraph{Planning.}
Due to the effective temporal modeling and advanced future prediction capabilities, \algname significantly improves UniAD by reducing its average collision rate within 3 seconds by $\sim$15\%. Moreover, it achieves a substantial decrease in the average planning displacement error by 0.21m and enables UniAD to outperform the state-of-the-art method, VAD \cite{jiang2023vad}, on nuScenes open-loop planning evaluation.
These improvements demonstrate the effectiveness of \algname as a valuable pre-training approach for end-to-end autonomous driving. The enhanced performance in collision avoidance and planning accuracy highlights the potential of \algname in enhancing the safety and efficiency of downstream autonomous driving applications.

\begin{table*}[t!]
  % \vspace{-.1in}
	\begin{center}
		\resizebox{\textwidth}{!}{
			\begin{tabular}{l|cc|ccc|cc|ccc|cccc|cc}
				\toprule[1.5pt]
				\multirow{2}{*}{Method} &
				\multicolumn{2}{c|}{Detection} & 
				\multicolumn{3}{c|}{Tracking} & 
				\multicolumn{2}{c|}{Mapping} & 
				\multicolumn{3}{c|}{Motion Forecasting} & 
				\multicolumn{4}{c|}{Future Occupancy Prediction} & 
                    \multicolumn{2}{c}{Planning}  \\
				& NDS $\uparrow$ & mAP $\uparrow$ & AMOTA$\uparrow$ & AMOTP$\downarrow$ &IDS$\downarrow$ & IoU-lane$\uparrow$ & IoU-road$\uparrow$ & minADE$\downarrow$ & minFDE$\downarrow$ & MR$\downarrow$ & IoU-n.$\uparrow$& IoU-f.$\uparrow$ & VPQ-n.$\uparrow$ & VPQ-f.$\uparrow$& avg.L2$\downarrow$  & avg.Col.$\downarrow$   \\
				\midrule
                    UniAD & 49.36 & 37.96 & 38.3 & 1.32 & 1054 & 31.3 & 69.1 & 0.75 & 1.08 & 0.158 & 62.8 & 40.1 & 54.6 & 33.9 & 1.12 & 0.27 \\
                    \algname 
& \bf 52.57 & \bf 42.33 & \bf 42.0 & \bf 1.25 & \bf 991 & \bf 33.2 & \bf 71.4 & \bf 0.67 & \bf 0.99 & \bf 0.149 & \bf 65.4 & \bf 42.1 & \bf 57.3 & \bf 36.4 & \bf 0.91 & \bf 0.23 \\
				\bottomrule[1.5pt]
			\end{tabular}
		}
	\end{center}
 \vspace{-15pt}
	\caption{
 \textbf{Performance gain of \algname for joint perception, prediction, and planning.} \algname consistently improves UniAD \cite{hu2023_uniad} on all tasks towards end-to-end autonomous driving, validating its effectiveness for scalable visual autonomous driving.
 }
	\label{tab:uniad}
\end{table*}
\paragraph{Joint Perception-Prediction-Planning.}
Finally, we summarize the improvements of \algname on the state-of-the-art end-to-end visual autonomous driving system, UniAD \cite{hu2023_uniad}, for joint perception, prediction, and planning. As depicted in \Cref{tab:uniad}, \algname brings substantial improvements on all sub-modules of UniAD for perception (Detection, Tracking, Mapping), prediction (Motion Forecasting and Future Occupancy Prediction), and planning at the same time.
These consistent improvements illustrate that visual point cloud forecasting effectively exploits the information of semantics, 3D geometry, and temporal dynamics behind the easily obtainable Image-LiDAR sequences. This, consequently, enables scalable visual autonomous driving.

\subsection{Ablative Study}
We conduct further analysis of \algname on improving the performance of downstream models with limited supervised data, and the effect of the \render operation on learning 3D geometric latent space. More ablation studies can be found in the supplementary materials.

\begin{figure}[t]
  \centering
  \vspace{-.15in}
  \includegraphics[width=0.95\linewidth]{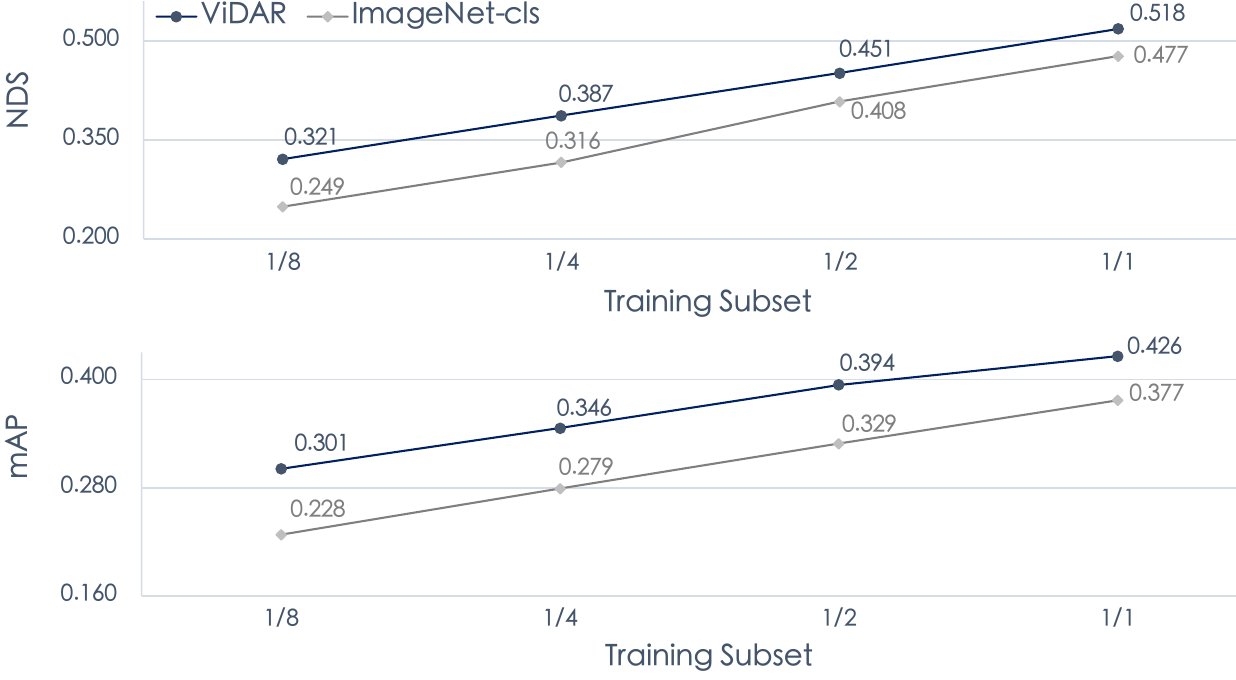}\\
  \vspace{-0.1in}
  \caption{\textbf{Validation of \algname on Fine-tuning with limited supervised data.} We verify \algname on supervision efficiency by reducing available annotations for 3D object detection during downstream fine-tuning (from the full training set to a 1/8th subset) and observe a continuous improvement on each subset.
  }
  \label{fig:data_scaling}
  \vspace{-0.1in}
\end{figure}
\paragraph{Efficiency of Supervised Pre-training.}
The primary objective of pre-training is to minimize the dependence on precise 3D annotations. In \Cref{fig:data_scaling}, we demonstrate the effectiveness of \algname in reducing the reliance of modern 3D detectors on accurate 3D box annotations. We fine-tune BEVFormer-base using partial 3D annotations on nuScenes, ranging from the full dataset to a $1/8$ subset.

As depicted in \Cref{fig:data_scaling}, \algname exhibits a remarkable reduction in the dependence on 3D annotations. Notably, BEVFormer, pre-trained by \algname, surpasses its counterpart under full supervision by 1.7\% mAP, while using only half of the supervised samples, \ie, 39.4\% mAP \vs 37.7\% mAP. Consequently, through \algname, we can reduce half of 3D annotations without sacrificing precision. Additionally, we observe a consistent trend of increasing improvements as the available supervision decreases. For instance, the mAP improvements are 4.9\%, 6.5\%, 6.7\%, and 7.3\% when fine-tuned on the full, half, a quarter, and 1/8th subsets. These results highlight the potential of \algname in harnessing large amounts of Image-LiDAR sequences.

\begin{table}[t]
  \centering
  \vspace{-.15in}
  \setlength{\tabcolsep}{3mm}
  \scalebox{0.8}{
  \begin{tabular}{l|cccccc}
    \toprule[1.5pt]
    \makecell[l]{Groups of \\
    Latent R.} & N/A & 1 & 2 & 4 & 8 & 16 \\
     \midrule
     NDS (\%) & 40.20 & 39.18 & 43.36 & 45.53 & 47.01 & \bf 47.58 \\
    \bottomrule[1.5pt]
  \end{tabular}
  }
  \vspace{-.1in}
  \caption{\textbf{Ablation of \render for downstream fine-tuning.} We compare the performance of 3D detection pre-trained by \algname without the \render operation (denoted as ``N/A'') and by \algname with different groups of \render.
  }
  \vspace{-.1in}
  \label{tab:latent_render}
\end{table}
\paragraph{Effect of \render operator.}
\render is the key component of \algname, which enables visual point cloud forecasting to effectively contribute to downstream applications. It addresses the ray-shaped features issue encountered during pre-training. In \Cref{tab:latent_render}, we verify its effectiveness by comparing the performance of downstream models pre-trained by \algname with the \render or not. The downstream model is BEVFormer-small \cite{li2022bevformer} for 3D object detection.  For context, the baseline performance with ImageNet-cls pre-training is 44.11\% NDS.

As depicted in \Cref{tab:latent_render}, when the \render is missing (denoted as ``N/A''), also referred to as the baseline in \Cref{sec:sec3.2}, a significant decline is observed in downstream performance after fine-tuning, from 44.11\% NDS to 40.20\% NDS. In contrast, with the 16-group \render, the performance improves to 47.58\% NDS, a notable 3.47\% NDS improvement over the baseline.

We conduct a comparison of \render with different parallel groups in \Cref{tab:latent_render} as well. The results demonstrate a consistent improvement by dividing channels into more groups and integrating information separately.

\section{Conclusion}
In this paper, we introduced visual point cloud forecasting, which predicts future point clouds from historical visual images, as a new pre-training task for end-to-end autonomous driving.
We developed \algname, a general model to pre-train visual BEV encoders, and designed a \render operator to solve the ray-shaped feature issue. To conclude, our work demonstrates that visual point cloud forecasting enables scalable autonomous driving.

\myparagraph{Limitations and Future Work.}
Though with the potential of scalability, in this paper, we mainly conduct pre-training on Image-LiDAR sequences from nuScenes dataset, of which the data scale is still limited. As for the future, we plan to scale up the pre-training data of \algname, study visual point cloud forecasting across diverse datasets, and use publicly available Image-LiDAR sequences as much as possible to train a foundation visual autonomous driving model~\cite{li2023opensourced}.

\section*{Acknowledgement}
OpenDriveLab is the autonomous driving team affiliated with Shanghai AI Lab.
This work was supported by National Key R\&D Program of China (2022ZD0160104), NSFC (62206172), and Shanghai Committee of Science and Technology (23YF1462000). We thank team members from OpenDriveLab for valuable feedback along the project.

{
    \small
    \bibliographystyle{ieeenat_fullname}
    \bibliography{main}
}

\newcommand\DoToC{%
  \startcontents
  {
      \hypersetup{linkcolor=black}
      \printcontents{}{1}{\noindent\textbf{\Large Appendix}\vskip5pt}
  }
}
% \maketitlesupplementary
\appendix
\onecolumn
\vspace{-.05in}
\maketitle
\DoToC

\bigskip

\vspace{-0.15in}
\section{Discussions}

Towards a better understanding of our work, we supplement intuitive questions
that one might raise.

\medskip

\myparagraph{Q1:} \textit{What is the relationship between \algname and world models?}

\smallskip

In general, \algname could be deemed as a world model - predicting the future world conditioned on observations and actions. However, it distinguishes itself from existing world models for autonomous driving \cite{zhang2023learning,wang2023drivedreamer,li2023drivingdiffusion,hu2023gaia1}. Unlike these models, which operate within the same modality for both inputs and outputs (\eg, image-in \& image-out or LiDAR-in \& LiDAR-out), \algname for the first time bridges different modalities. It leverages historical visual sequences as inputs to predict future point clouds. By utilizing \algname, it becomes possible to generate various future point clouds, conditioned on different future ego-vehicle motions, using visual inputs. This holds significant potential in training vision autonomy in 3D.

\vspace{0.05in}
\myparagraph{Q2:} \textit{Why use point clouds (LiDAR) as outputs, instead of future images?}

\smallskip
Compared to images, point clouds offer a highly precise depiction of the 3D environment, effectively capturing scene structure, object positions, and geometric properties. This detailed representation proves advantages in various 3D tasks, including perception, reconstruction, and rendering. Consequently, in \algname, we opt to employ point clouds as the prediction target. This choice enables the model to extract 3D geometry from visual inputs, thereby empowering downstream models.

\vspace{0.05in}
\myparagraph{Q3:} \textit{What are potential applications and future directions of \algname?}

\smallskip
In our work, we have demonstrated the effectiveness of \algname as a pre-training approach for enhancing downstream end-to-end autonomous driving models~\cite{hu2023_uniad, chen2023e2esurvey}. Additionally, considering its capability as a world model, there is significant promise in utilizing \algname as a simulator for model-based reinforcement learning, thereby bolstering the decision-making abilities of vision-based autonomous agents. The application of \algname in this context opens up avenues for future research. For instance, to facilitate its scalability, investigations into the utilization of Image-LiDAR sequences from diverse datasets are necessary. Furthermore, exploring the combination of \algname with other single-modality world models to create more favorable data simulations presents intriguing prospects for future inquiry.

\section{Implementation Details}
\begin{figure*}[h]
        \vspace{-.1in}
	\centering
	\includegraphics[width=.9\linewidth]{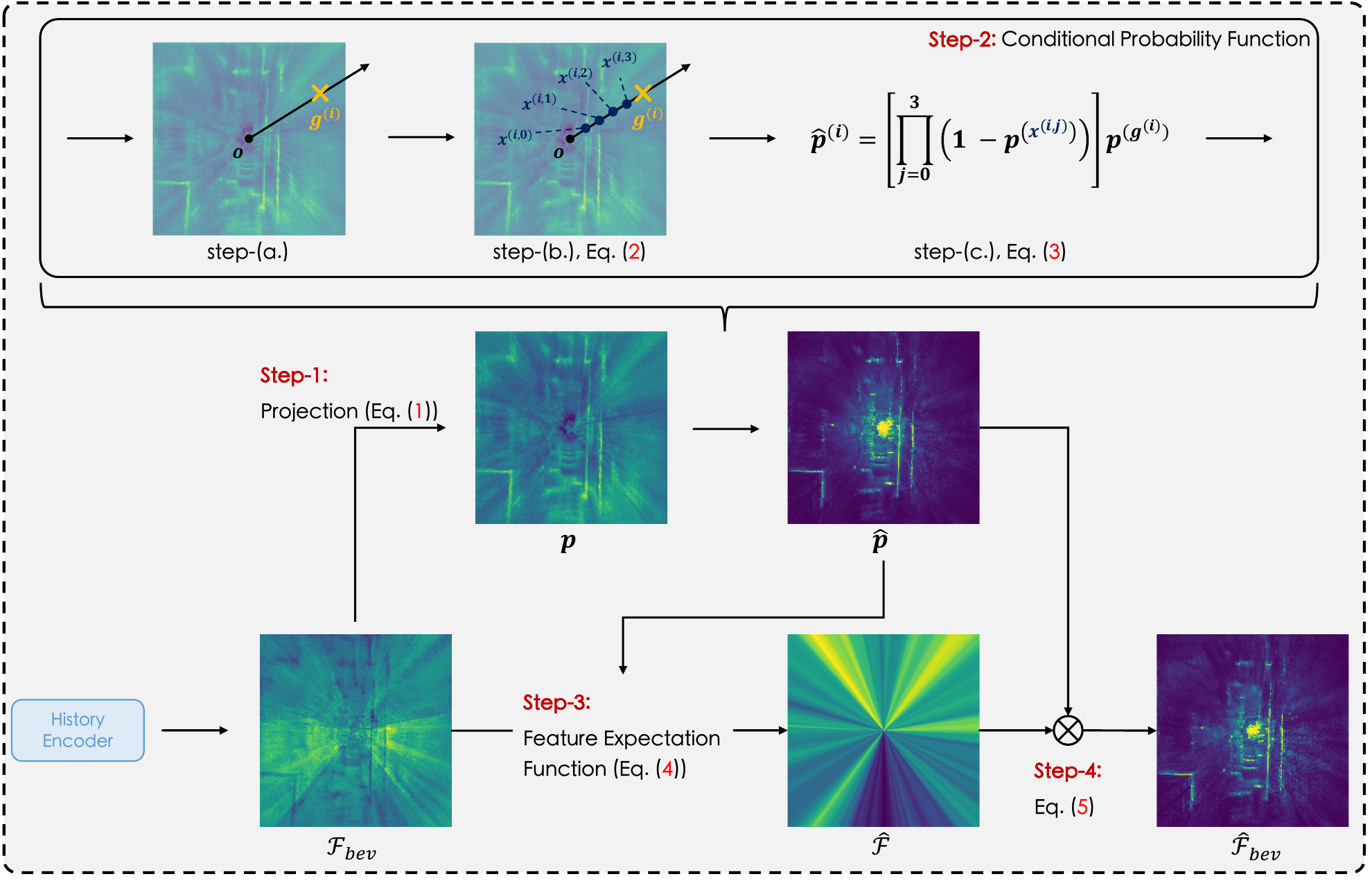}
        \vspace{-.1in}
	\caption{\textbf{Detailed architecture of the \render operator.} We only show the case of single-group for simplicity. Given the visual BEV embedding $\mathcal{F}_{\text{bev}}$ from the \encoder, the \render operator performs a series of steps. First, it employs a projection layer to estimate the independent probability map, $\mathbf{p}$. Subsequently, the conditional probability function generates the conditional probability of each BEV grid, denoted as $\hat{\mathbf{p}}^{(i)}$. Then, we compute ray-wise features using a feature expectation function, and finally, the resulting ray-wise features are multiplied with the conditional probability map to yield the geometric feature space.
	}
	\label{fig:supp_latent_render}
\end{figure*}

\myparagraph{\render.}
We now delve into the specific implementation details of the \render operator. As shown in \Cref{fig:supp_latent_render}, the \render operator comprises a total of 4 steps. In the first step, given the BEV embeddings from the \render, we use a projection layer with output channels of $G$ to estimate independent probability maps:
\vspace{-.05in}
\begin{equation} \label{eq:lr_projection}
	\begin{aligned}
            \mathbf{p} = \texttt{Projection}(\mathcal{F}_{\text{bev}}),
	\end{aligned}
\vspace{-.05in}
\end{equation}
where $G$ is the group number for multi-group \render. To simplify, we focus on where $G=1$. In cases of multiple groups, we can easily divide $\mathcal{F}_{\text{bev}}$ into $G$ parts along the channels, and apply \render on each part accordingly.

In the second step, we compute the conditional probability of each BEV grid along its respective ray. As shown in step-(a.) of \Cref{fig:supp_latent_render}, for the $i$-th BEV grid with coordinates $\mathbf{g}^{(i)}=\{x_i, y_i \}$, we begin by casting a ray from the origin point $\mathbf{o}$, typically the center of BEV feature maps, towards the target BEV grid. The direction of the ray is determined as: $\mathbf{d}^{(i)} = (\mathbf{g}^{(i)} - \mathbf{o}) / (||\mathbf{g}^{(i)} - \mathbf{o}||_2)$. Subsequently, we collect a set of prior waypoints along the ray, which are closer to the origin point in distance compared to BEV grid $i$. These waypoints are selected at uniform distance intervals, termed as $\lambda$, as:
\vspace{-.05in}
\begin{equation} \label{eq:lr_waypoints}
	\begin{aligned}
            \mathbf{x}^{(i, j)} = \mathbf{o} + j \lambda \mathbf{d}^{(i)},\ \text{where}\ j\lambda < ||\mathbf{g}^{(i)} - \mathbf{o}||_2.
	\end{aligned}
% \vspace{-.05in}
\end{equation}
Here, $j$ represents the index of different waypoints, and the condition $j\lambda < ||\mathbf{g}^{(i)} - \mathbf{o}||_2$ ensures that the distance of these waypoints from the origin point is smaller than that of the corresponding BEV grid. An example is depicted in step-(b.) of \Cref{fig:supp_latent_render}, where we collect 4 prior waypoints with coordinates as $\mathbf{x}^{(i,j)}$ where $j \in \{0,1,2,3\}$ for the $i$-th BEV grid. Moving on to step-(c.) of \Cref{fig:supp_latent_render}, we compute the conditional probability of the target BEV grid based on the independent probability of its prior waypoints. In general, the conditional probability function is defined as:
\begin{equation} \label{eq:lr_conditional}
	\begin{aligned}
            \hat{\mathbf{p}}^{(i)} = 
                \bigg [
                    \prod_j(1 - \mathbf{p}^{(\mathbf{x}^{(i,j)})})
                \bigg ] \mathbf{p}^{(\mathbf{g}^{(i)})},
	\end{aligned}
\end{equation}
where $\mathbf{p}^{(:)}$ is the bilinear interpolation function to compute the corresponding probability of the position $(:)$ from $\mathbf{p}$.

In the third step, we compute the ray-wise features according to the BEV embedding and conditional probability map:
\begin{equation} \label{eq:lr_expectation}
	\begin{aligned}
            \hat{\mathcal{F}}^{(i)} = 
                    \sum_k \hat{\mathbf{p}}^{(i, k)} \mathcal{F}_{\text{bev}}^{(k)},\ \text{where}\ \mathbf{d}^{(k)} = \mathbf{d}^{(i)},
	\end{aligned}
\end{equation}
where $k$ indexes those BEV grids in the same direction as the $i$-th BEV grid. 

As illustrated in \Cref{fig:supp_latent_render}, after the feature expectation function, all BEV grids lying on the same ray share the same global feature embeddings. Therefore, finally, in order to highlight those BEV grids with higher conditional probability, we multiply the ray-wise features and the conditional probability map to obtain the geometric feature responses:
\begin{equation} \label{eq:lr_multiply}
	\begin{aligned}
            \hat{\mathcal{F}}_{\text{bev}} = 
                    \hat{\mathbf{p}} \cdot  \hat{\mathcal{F}}.
	\end{aligned}
\end{equation}

In our concrete implementation, we set $G$ and $\lambda$ as 16 and 1, respectively.

\section{Additional Ablative Studies}

\subsection{Structure of \render}

\begin{wraptable}{r}{0.4\textwidth}
    \vspace{-0.2in}
    \setlength{\tabcolsep}{3.5mm}
    \centering\small
    \begin{tabular}{c|c|c}
    \toprule[1.5pt]
         \makecell[l]{Cond. Prob. \\ Func., Eq. \eqref{eq:lr_conditional}} & \makecell[l]{Feat. Exp. \\ Func., Eq. \eqref{eq:lr_expectation}} & NDS (\%) \\ 
         \midrule
        - & - & 40.20 \\
        \midrule
        $\surd$ & - & 47.34 \\
        \midrule
        $\surd$ & $\surd$ & \bf47.58 \\
    \bottomrule[1.5pt]
    \end{tabular}
    \vspace{-.1in}
    \caption{\textbf{Ablation on each component of \render design.} ``Cond. Prob. Func.'' and ``Feat. Exp. Func.'' represent the conditional probability function and the feature expectation function, respectively.}
    \vspace{-0.1in}
    \label{tab:abla_lr_structure}
\end{wraptable}
In \Cref{tab:abla_lr_structure}, we investigate the effectiveness of conditional probability function (Eq. \eqref{eq:lr_conditional}, Step-2 in \Cref{fig:supp_latent_render}) and feature expectation function (Eq. \eqref{eq:lr_expectation}, Step-3 in \Cref{fig:supp_latent_render}). For context,  we use BEVFormer-small \cite{li2022bevformer} with ImageNet-cls pre-training as the baseline. We evaluate different functions by first pre-training the baseline using \algname under various structures and then comparing the downstream 3D detection performance. The baseline result, 40.20\% NDS, is obtained by pre-training the model using the straightforward pipeline, mentioned in \Cref{sec:sec3.2} of the main paper, which comprises the \encoder, \decoder, and differentiable ray-casting.

To evaluate the effectiveness of the conditional probability function solely, we directly multiply the BEV embeddings $\mathcal{F}_{\text{bev}}$ with the aggregated conditional probability map $\hat{\mathbf{p}}$ to obtain the features for future point cloud prediction. As illustrated in the second row of \Cref{tab:abla_lr_structure}, the conditional probability function greatly alleviates the ray-shaped feature issue, and brings a $7.14\%$ NDS improvement compared to the simple baseline. Then, by introducing the feature expectation function, the \render operator integrates the entire ray-wise feature, which further brings downstream performance improvements.

\begin{wraptable}{r}{0.4\textwidth}
    \vspace{-0.2in}
    \setlength{\tabcolsep}{3.0mm}
    \centering\small
    \begin{tabular}{c|c|c}
    \toprule[1.5pt]
         \makecell[l]{No. of Future \\ Supervision} & 
         \makecell[c]{Detection \\ NDS (\%) $\uparrow$} &
         \makecell[c]{Tracking \\ AMOTA (\%) $\uparrow$}\\ 
         \midrule
        0 & 54.04 & 42.12 \\
         \midrule
        1 & 53.73 & 43.79 \\
         \midrule
        3 & 53.60 & 44.74 \\
         \midrule
        6 & 53.81 & \bf 45.10 \\
    \bottomrule[1.5pt]
    \end{tabular}
    \vspace{-.1in}
    \caption{\textbf{Ablation on future supervision.} More future supervisions during the pre-training stage bring consistent improvements in tracking performance.}
    \vspace{-.1in}
    \label{tab:abla_future}
\end{wraptable}
\subsection{Future Predictions}
In this study, we aim to showcase the benefits of future prediction in downstream tasks that involve temporal modeling. We conduct ablation studies on the task of Multi-Object Tracking, as it necessitates the model to associate 3D objects across different frames, thereby reflecting its temporal modeling capabilities. For our experiments, we use the first stage of UniAD \cite{hu2023_uniad} as the tracking model, and assess its performance on downstream tracking after being pre-trained with \algname using varying frames of future point cloud supervision.

As shown in \Cref{tab:abla_future}, we evaluate four settings which utilize ``0'', ``1'', ``3'', and ``6'' future frames to supervise \algname when visual point cloud forecasting pre-training. The setting labeled as ``0  future supervision'' implies that we solely utilize \algname to reconstruct LiDAR point clouds for the current frame, without any future prediction and supervision. As illustrated, we observe consistent improvements in tracking performance when using more future frames for supervision. This observation highlights the beneficial impact of visual point cloud forecasting on the temporal modeling capabilities of downstream models.

\subsection{Pre-train Structure}

\begin{wraptable}{r}{0.5\textwidth}
    \vspace{-0.2in}
    \setlength{\tabcolsep}{3.0mm}
    \centering\small
    \begin{tabular}{c|c|c|c}
    \toprule[1.5pt]
         2D Backbone & FPN-neck & \makecell[l]{View-Transform} & NDS (\%) \\ 
         \midrule
        - & - & - & 44.11 \\
        \midrule
        $\surd$ & $\surd$ & - & 44.74 \\
        \midrule
        $\surd$ & $\surd$ & $\surd$ & 47.58 \\
    \bottomrule[1.5pt]
    \end{tabular}
    \vspace{-0.1in}
    \caption{\textbf{Ablation on pre-training components.} \algname mostly benefits the view-transformation part of visual BEV encoders.}
    \label{tab:abla_pretrain_structure}
\end{wraptable}
In \Cref{tab:abla_pretrain_structure}, we investigate the effect of \algname pre-training on different components of downstream BEV encoders. We achieve this by loading different sets of pre-trained model parameters during the downstream fine-tuning process. For context, in this ablation study, we use BEVFormer-small as the downstream model for 3D detection. The baseline performance is 44.11\% NDS.

As illustrated, the improvements primarily stem from the pre-training of the view-transformation component, which is responsible for extracting BEV features from perspective multi-view image features. It is reasonable considering that the view-transformation module plays a crucial role in correlating 2D features with 3D geometries and scene structures, as discussed in previous works~\cite{li2023bevdepth}. Furthermore, this highlights the compatibility of \algname pre-training with any advanced 2D image pre-training techniques, which could lead to consistent performance improvements when combined with more advanced image backbones, as demonstrated in Table \textcolor{red}{3} and Table \textcolor{red}{4} of the main paper.

\section{Qualitative Results}

\subsection{\render}
\begin{figure*}[h]
	\centering
	\includegraphics[width=1.0\linewidth]{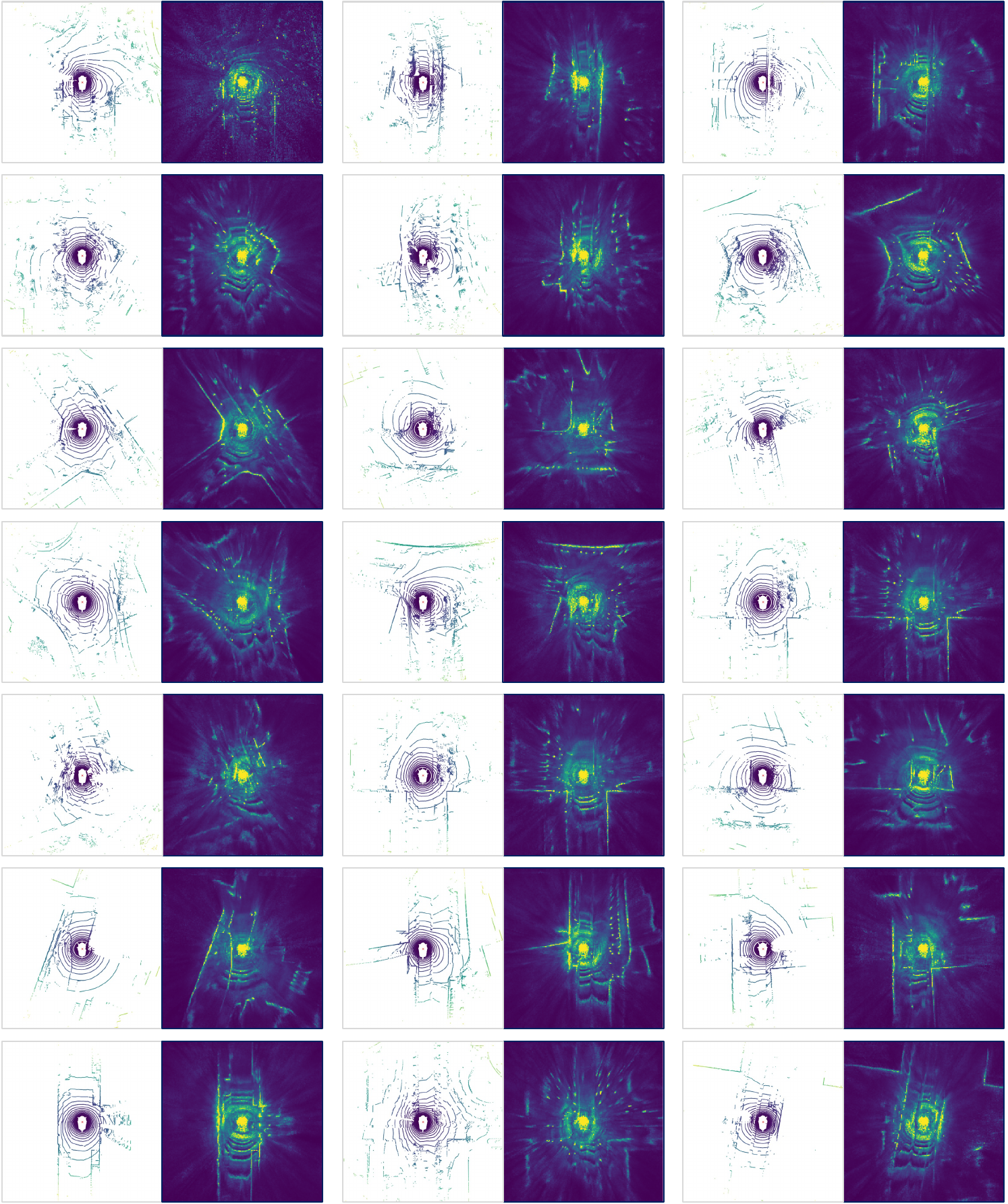}
	\vspace{-0.1in}
	\caption{\textbf{Visualization of geometric features from the \render operator.} In each pair, we present the ground-truth LiDAR point clouds on the left and the BEV features produced by \algname from multi-view image inputs on the right. It is evident that, utilizing the \render operator, \algname successfully captures the underlying 3D geometry in the latent space, resulting in precise descriptions of the 3D world through feature responses.
	}
	\label{fig:supp_latent_render_feature}
\end{figure*}

\Cref{fig:supp_latent_render_feature} presents the effectiveness of the \render operator in formulating geometric features from visual sequence inputs. Each pair of images in \Cref{fig:supp_latent_render_feature} displays the ground-truth point cloud on the left and visualizes the features, denoted as $\hat{\mathcal{F}}_{\text{bev}}$, on the right. As depicted, \algname successfully captures the 3D geometry from multi-view visual sequences and effectively extracts geometric features that accurately represent the underlying 3D world in the latent space.

\begin{figure*}[h]
	\centering
	\includegraphics[width=1.0\linewidth]{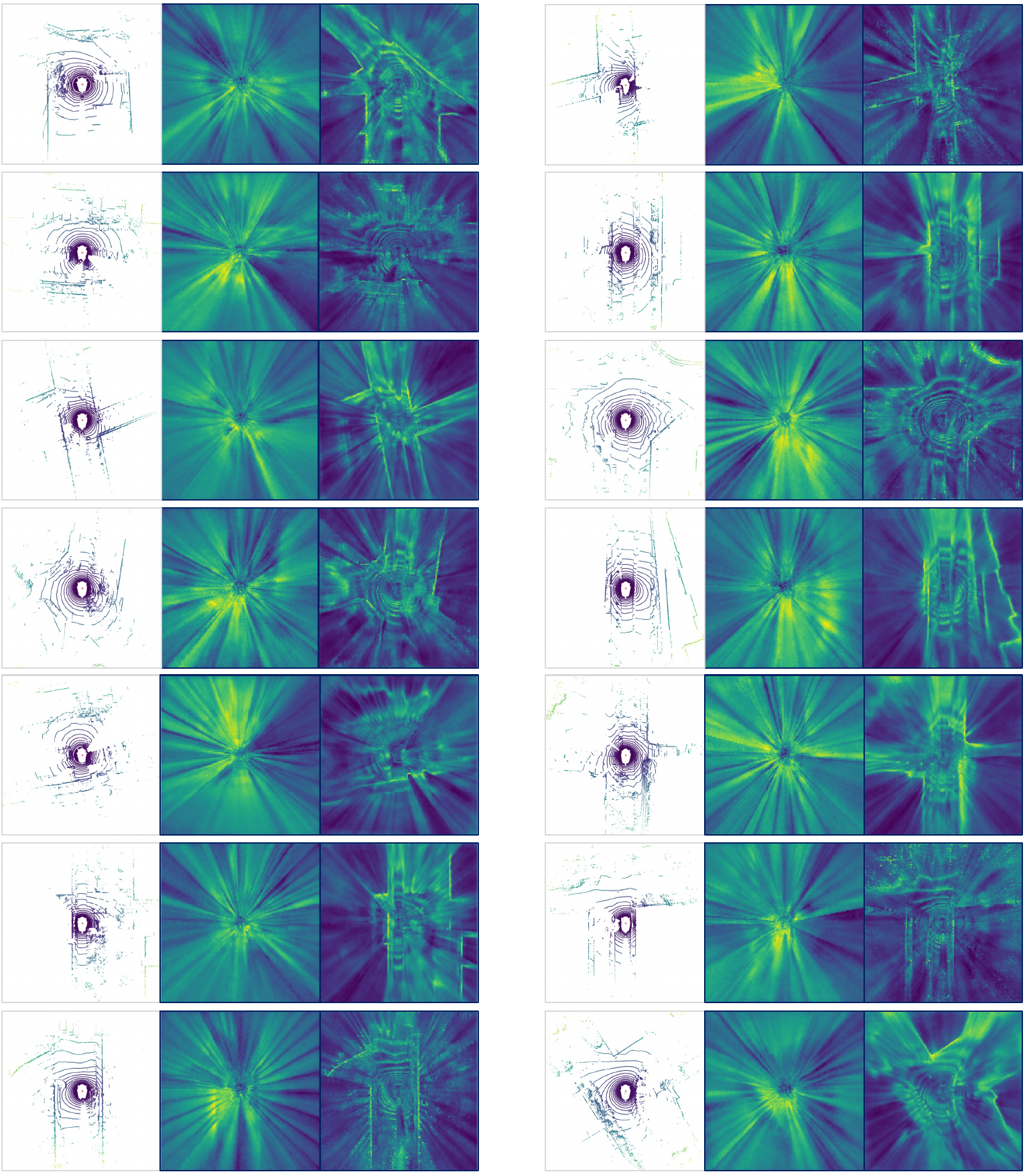}
	\vspace{-0.1in}
	\caption{\textbf{Visualization of BEV features from visual BEV encoder pre-trained by the differentiable ray-casting baseline and \algname with \render operator.} In each triplet, we present the ground-truth point cloud on the left, the features pre-trained by the differentiable ray-casting baseline in the middle, and the features pre-trained by \algname on the right. As illustrated, the integration of the \render operator in \algname proves advantageous as it successfully mitigates the occurrence of ray-shaped features during visual point cloud forecasting pre-training. Consequently, \algname empowers the BEV encoders to extract informative and discriminative BEV features from visual sequence inputs.
	}
	\label{fig:supp_pretrained_feature}
\end{figure*}

Next, in \Cref{fig:supp_pretrained_feature}, we compare the BEV features, 
$\mathcal{F}_{\text{bev}}$, extracted by BEV encoders pre-trained using the 
differentiable ray-casting baseline (depicted in the middle) and our 
\algname (depicted on the right). As illustrated, the ray-casting baseline encounters the issue of ray-shaped features after pre-training. In contrast, our \algname, thanks to the inclusion of \render during the pre-training procedure, effectively highlights the responses of BEV grids preserving geometric information. Consequently, it extracts discriminative features after pre-training, which, in turn, benefits downstream fine-tuning.

\subsection{Visual Point Cloud Forecasting}
\begin{figure*}[h]
	\centering
        \vspace{-.2in}
	\includegraphics[width=0.99\linewidth]{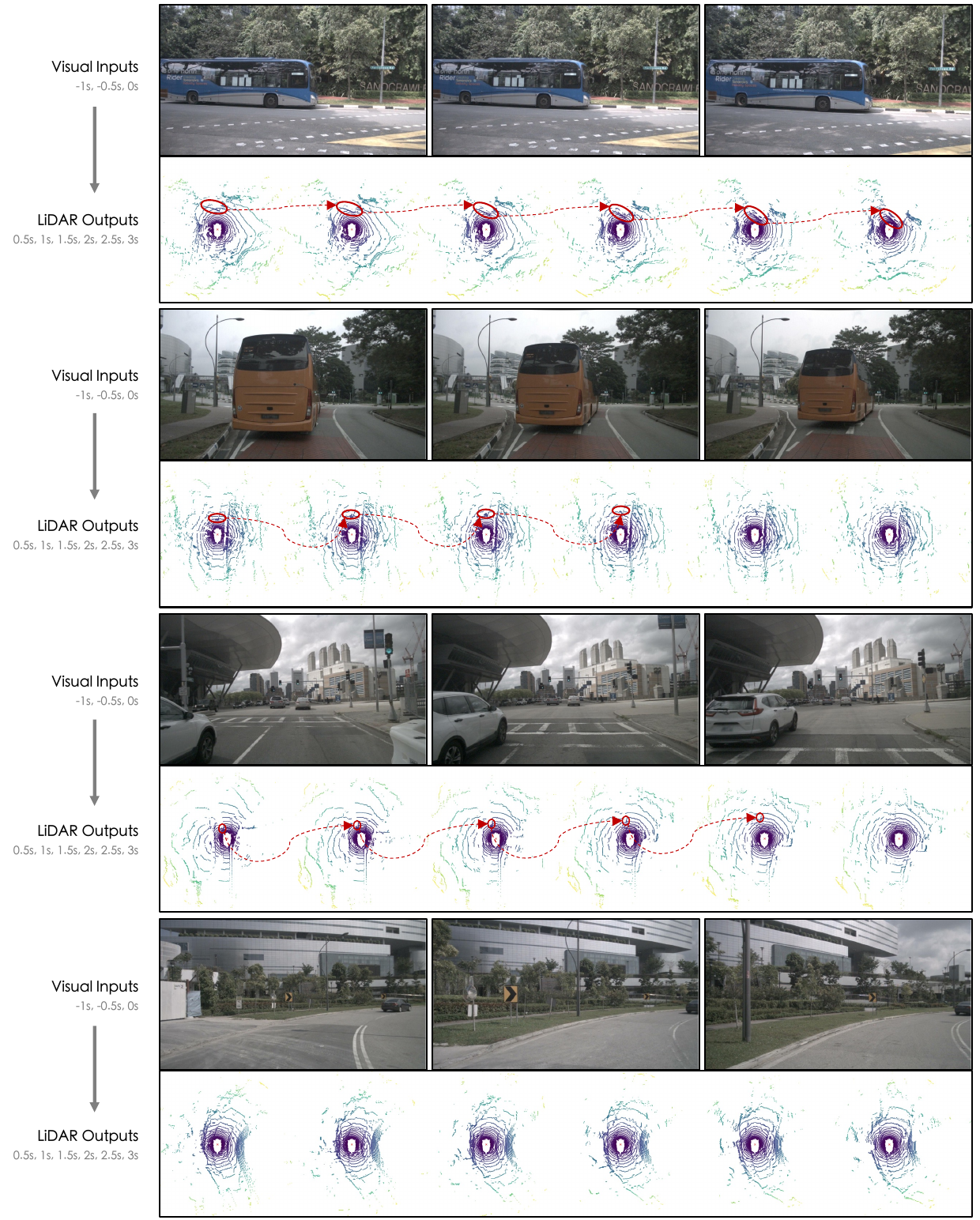}
 \vspace{-.1in}
	\caption{\textbf{Qualitative results of \algname for visual point cloud forecasting on nuScenes validation set.} Top: historical visual inputs in 1 second; bottom: future point cloud predictions in 3 seconds. The first, second, and third rows provide examples of modeling relative motions between the ego-vehicle and other parked or moving objects (highlighted in \textcolor{red}{\bf red circles}); the fourth row shows the future estimation where the ego-vehicle is turning right. As illustrated, \algname effectively captures the information of 3D geometry and temporal dynamics, and thus correctly estimates future point clouds from visual sequence inputs. All point cloud visualizations are under the ego coordinates.
	}
	\label{fig:supp_pc_forecast}
\end{figure*}

In \Cref{fig:supp_pc_forecast}, we provide visual examples of \algname for predicting future point clouds based on historical visual images. The historical visual inputs captured within a 1-second timeframe are displayed in the upper portion, while the corresponding predicted future point clouds spanning 3 seconds are shown in the lower portion.

In the first row, we present an example of the ego-vehicle executing a left turn. As observed, \algname adeptly captures the related position and orientation between the ego-vehicle and the parked blue bus in its future predictions. Moving on to the second and third rows, we illustrate instances where \algname successfully captures the relative motions between the ego-vehicle and other moving objects, such as the yellow bus in the second row and the white car in the third row. By analyzing the LiDAR outputs, \algname accurately understands that moving objects exhibit faster movement compared to the ego-vehicle. Consequently, it estimates their positions with increasing relative distances as time progresses. Then, the fourth row showcases an example of the ego-vehicle executing a right turn. This case demonstrates \algname's effective modeling of the road map based on the historical visual sequences. It is important to note that all LiDAR point cloud visualizations are presented within the coordinate space of the ego-vehicle, with the ego-vehicle situated at the center of the 3D space.

\begin{figure*}[h]
	\centering
	\includegraphics[width=1.0\linewidth]{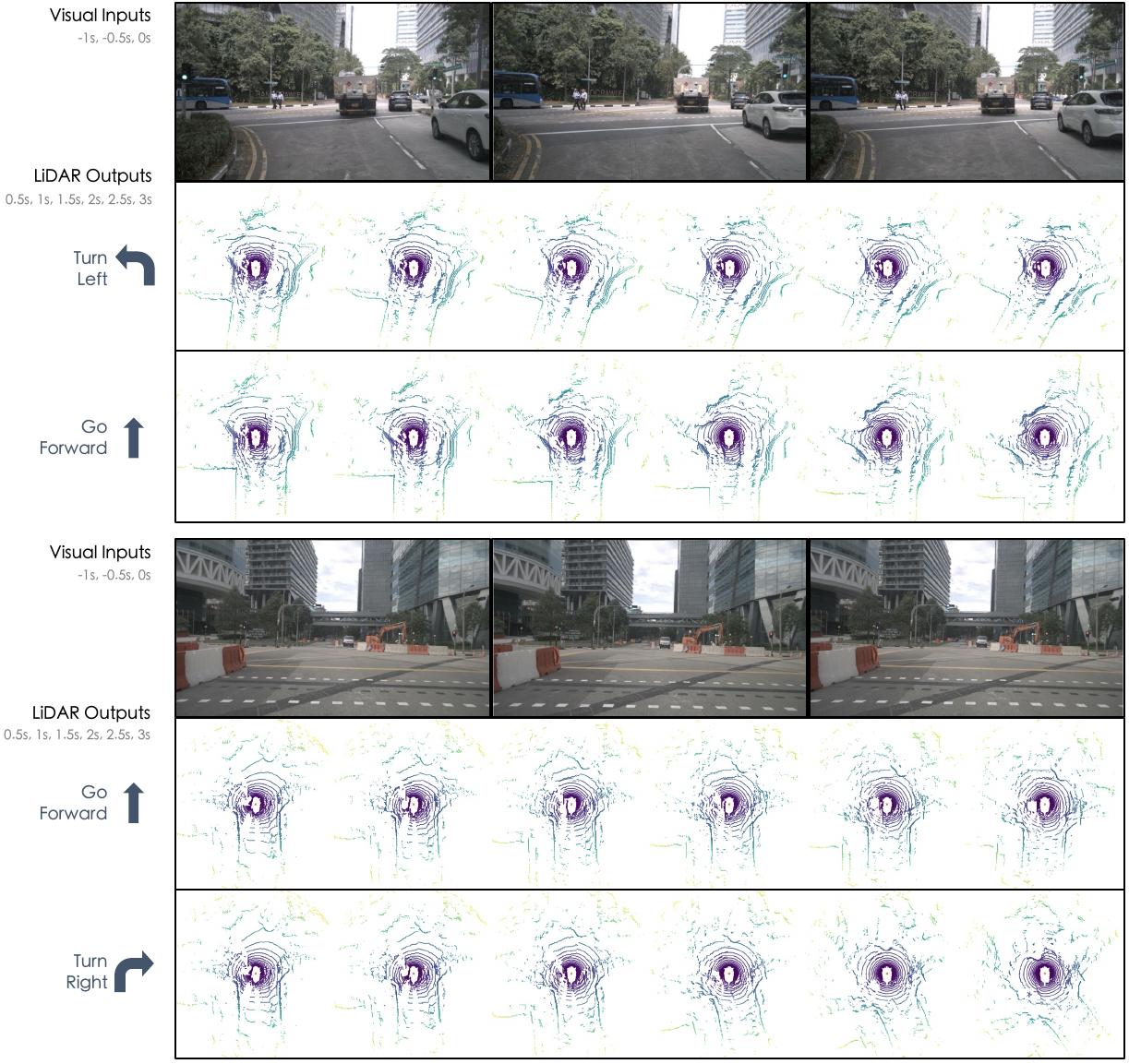}
	\caption{\textbf{Qualitative results of \algname conditioned on different future ego-vehicle motion conditions.} \algname can simulate different future point clouds based on the specific future ego-vehicle conditions. All point cloud visualizations are under the ego coordinates.
	}
	\label{fig:supp_vidar_control}
\end{figure*}

Additionally, in \Cref{fig:supp_vidar_control}, we demonstrate the capability of \algname to simulate various future point clouds based on specific ego-vehicle motions, such as turning left, going straight, and turning right. This showcases the potential of \algname as a visual world model for autonomous driving, where it utilizes visual image inputs to generate simulated future point clouds. Such simulations can be valuable for model-based reinforcement learning for training vision autonomy in an unsupervised manner.

\subsection{End-to-end Autonomous Driving}
\begin{figure*}[h]
	\centering
	\includegraphics[width=1.0\linewidth]{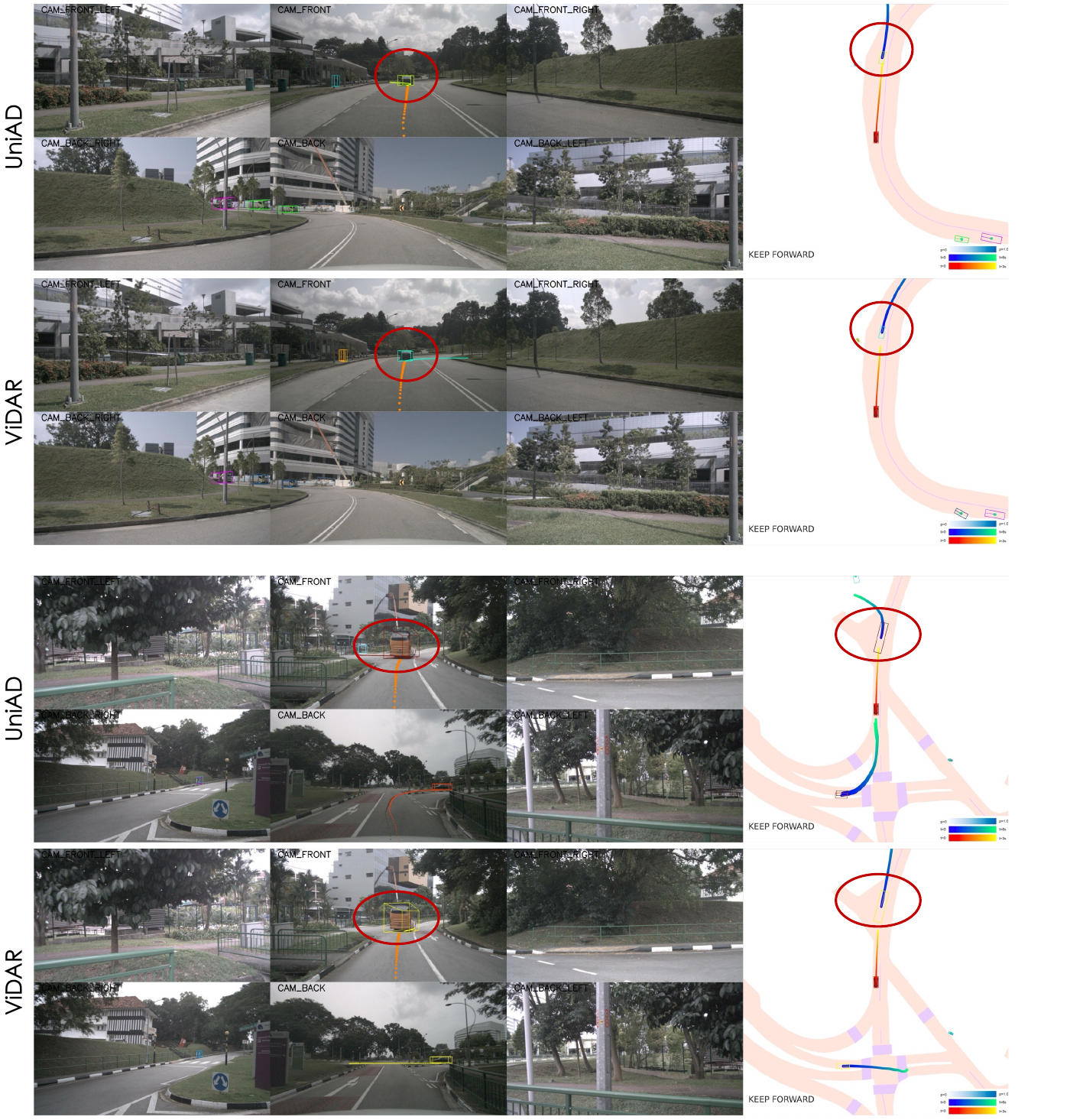}
	\caption{\textbf{Qualitative comparisons between UniAD with and without \algname pre-training.} \algname effectively models the temporal dynamics by estimating future points during pre-training, which in turn improves UniAD in predicting the future motion of moving objects.
	}
	\label{fig:supp_viz_uniad}
\end{figure*}

Lastly, in \Cref{fig:supp_viz_uniad}, we present a comparison between UniAD with and without \algname pre-training for end-to-end autonomous driving. As illustrated, the inclusion of \algname pre-training enables UniAD to generate more precise future trajectories for other moving objects (highlighted in \textcolor{red}{\bf red circles}). This improved prediction accuracy plays a crucial role in enhancing the planning process for safety-critical end-to-end autonomous driving scenarios.

\clearpage

\clearpage

\clearpage

\clearpage

\clearpage

\end{document}